\documentclass[10pt,twocolumn,letterpaper]{article}

\usepackage[pagenumbers]{cvpr} 

\usepackage[dvipsnames]{xcolor}

\definecolor{mygreen}{RGB}{29, 177, 0}

\def\qaego4d{\textsc{QaEgo4D}}
\def\ourdataset{\textsc{EgoTimeQA}}
\def\ourmethod{GroundVQA}

\usepackage{graphicx}
\usepackage{amsmath}
\usepackage{amssymb}
\usepackage{booktabs}
\usepackage{colortbl}
\usepackage{arydshln} 
\usepackage{multirow}
\usepackage{stfloats}
\usepackage{makecell}
\usepackage{tabu}
\usepackage{listings}
\usepackage{subcaption}
\usepackage[hang,flushmargin]{footmisc}

\usepackage[utf8]{inputenc}
\usepackage{pgfplots}
\DeclareUnicodeCharacter{2212}{−}
\usepgfplotslibrary{groupplots,dateplot}
\usetikzlibrary{patterns,shapes.arrows}
\pgfplotsset{compat=newest}

\definecolor{cvprblue}{rgb}{0.21,0.49,0.74}
\usepackage[pagebackref,breaklinks,colorlinks,citecolor=cvprblue]{hyperref}

\def\paperID{102} 
\def\confName{CVPR}
\def\confYear{2024}

\title{Grounded Question-Answering in Long Egocentric Videos}

\author{Shangzhe Di$^{1}$ \hspace{30pt} Weidi Xie$^{1,2}$\\[10pt]
$^{1}$CMIC, Shanghai Jiao Tong University, China
\hspace{20pt}
$^{2}$Shanghai AI Lab, China
}

\begin{document}

\maketitle
\begin{abstract}
Existing approaches to video understanding, mainly designed for short videos from a third-person perspective, are limited in their applicability in certain fields, such as robotics. In this paper, we delve into open-ended question-answering (QA) in long, egocentric videos, which allows individuals or robots to inquire about their own past visual experiences. This task presents unique challenges, including the complexity of temporally grounding queries within extensive video content, the high resource demands for precise data annotation, and the inherent difficulty of evaluating open-ended answers due to their ambiguous nature. Our proposed approach tackles these challenges by (i) integrating query grounding and answering within a unified model to reduce error propagation; (ii) employing large language models for efficient and scalable data synthesis; and (iii) introducing a close-ended QA task for evaluation, to manage answer ambiguity. Extensive experiments demonstrate the effectiveness of our method, which also achieves state-of-the-art performance on the QAEgo4D and Ego4D-NLQ benchmarks. Code, data, and models are open-sourced \footnote{\url{https://github.com/Becomebright/GroundVQA}}.
\end{abstract}    
\section{Introduction}
\label{sec:intro} 
In the literature, existing video perception tasks have primarily focused on videos in third-person view, for example, action recognition~\cite{kinetics, activitynet, something}, video-language grounding~\cite{tacos, activitynet-captions, charades-sta}, and video question-answering~\cite{activitynet-qa, how2qa, nextqa}, these videos are short, {\em e.g.}, typically ranging from 10 seconds to one minute.
Until recently, the proposal of Ego4D dataset~\cite{ego4d} re-ignites the interest of video understanding from egocentric views, where the inputs are normally long, continuous video streams from the first-person point of view, {\em i.e.}, seeing the world through the eyes of an agent actively engaged with its environment, 
which resembles an important step towards deploying vision models into real-world scenarios, such as robotics and augmented reality.

\begin{figure}[t]
\centerline{\includegraphics[width=\columnwidth]{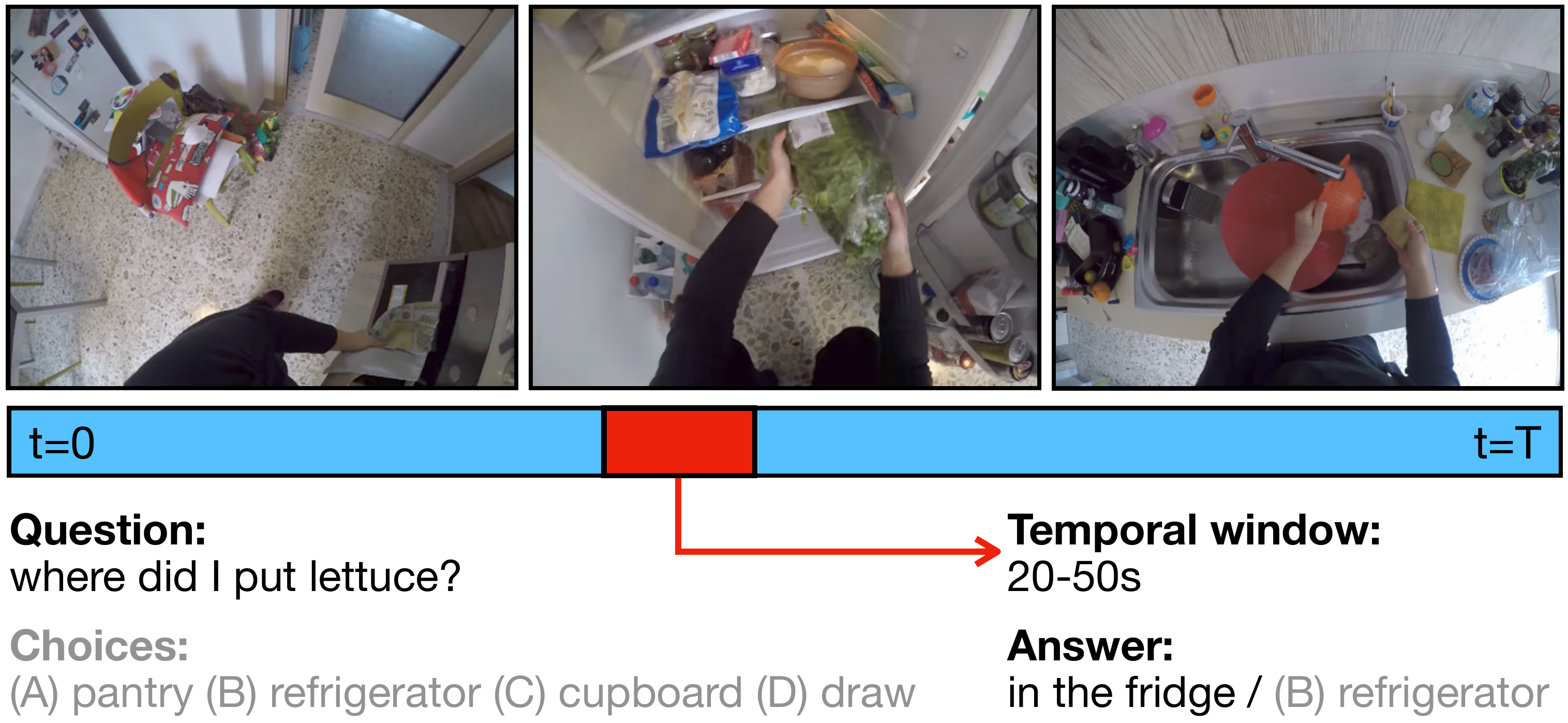}}
\vspace{-3pt}
\caption{We propose a unified model for addressing grounded question answering in long egocentric videos, {\em i.e.}, simultaneously identifying the temporal window to a question, generating answers in natural language~(OpenQA task), or \textcolor{gray}{picking answers from candidate choices~(CloseQA task)}.}\label{fig:teaser}
\vspace{-0.5cm}
\end{figure}

In this paper, we consider question answering~(QA) in long, egocentric videos, as illustrated in Fig.~\ref{fig:teaser}. Given questions about an egocentric video, {\em e.g.}, ``where did I put lettuce'', we aim to build a visual system that can answer the raised question in free-form language. This task serves two purposes: enhancing episodic memory~\cite{episodic_memory}, {\em i.e.}, allowing a person or robot to ask questions on the fly about their own past visual experience; or probing the multi-modal reasoning abilities of deep models.

Question-answering~(QA) in long egocentric videos is challenging, primarily due to the complexity of temporally grounding and generating answers to the queries within extensive video content. A pioneer work, overlooking the importance of query grounding, achieves unsatisfactory QA performance that merely outperforms ``blind guessing''~\cite{qaego4d}. On the other hand, research about temporal grounding in long egocentric videos, while achieving good progress, is limited in practical uses without the QA ability. A potential fix would be chaining models from the two areas, {\em i.e.}, starting by localizing the temporal window to which the question relates, and followed by answering based on the corresponding video context. However, such a method is often ineffective due to error propagation. To address these challenges, we propose to train a unified model for simultaneous query grounding and answering, as shown in Fig.~\ref{fig:teaser}. 
The unified training has three advantages: 
{\em First}, by training the grounding task, the model can better grasp query-relevant information from the long videos, which is helpful for effective QA; 
{\em Second}, simultaneously training these two tasks can reduce error accumulation thanks to the synergy effect~\cite{ruder2017overview} in deep models; 
{\em Third}, predicting a temporal window helps to understand the cause of failure. Thus, we propose to solve the query grounding and answering concurrently, namely \textbf{\ourmethod}.

Nevertheless, training the unified architecture demands significant resources and effort to manually annotate the triplets -- comprising a question, answer, and temporal window -- on lengthy videos. Limited training data poses a major challenge in training large models with millions of parameters. To combat this issue, we establish an automatic pipeline that leverages large language models (LLMs) to generate abundant training samples.
This pipeline prompts LLMs to transform the plentiful, timestamped narrations in Ego4D into QA pairs, and estimates corresponding temporal windows. As a result, we produce 303K data samples from 5,389 video clips, which is a 30-fold increase over the existing dataset~\cite{qaego4d}. Our newly created pre-training dataset, named \textbf{\ourdataset}, effectively mitigates overfitting and significantly enhances grounding and QA performance.

In addition, we face challenges in evaluating open-ended answers, 
{\em i.e.}, free-form language generation.
Although open-ended QA is more representative of real-world scenarios where users interact with systems in natural language, it is a common consensus that the existing metrics like BLEU~\cite{bleu}, METEOR~\cite{meteor}, and ROUGE~\cite{rouge} are not fully satisfactory. To address this, we introduce CloseQA, an alternative close-ended task, where the model is asked to pick the correct answer from a set of candidate choices. We again leverage LLMs to generate plausible but incorrect answers, providing training and testing data for CloseQA.

The rest of the paper is structured as follows:
Sec.~\ref{sec:related_work} summarizes and discusses the relevant literature. 
Sec.~\ref{sec:method} begins with an introduction to the proposed model for simultaneous query grounding and answering, followed by an automatic pipeline for augmenting the existing training dataset in a scalable manner. 
In Sec.~\ref{sec:experiments}, comprehensive ablation studies are presented to demonstrate the effectiveness of our proposed techniques. 
Consequently, our model achieves state-of-the-art performance on the \qaego4d~\cite{qaego4d} and Ego4D-NLQ~\cite{ego4d} benchmarks.
\section{Related Work}
\label{sec:related_work}

\vspace{2pt} \noindent \textbf{Video language grounding.}
Video language grounding (VLG), initially proposed by Hendricks et al.~\cite{anne2017localizing}, involves identifying and segmenting specific temporal intervals within third-person view videos based on a natural language description or query~\cite{glipv2,unloc,univtg}. Several datasets and benchmarks, such as Charades-STA\cite{charades-sta} and TACoS~\cite{tacos}, have been curated to support research in this field. Notably, the Ego4D-NLQ~\cite{ego4d} dataset features long-form egocentric videos paired with natural language queries. The NaQ dataset~\cite{naq} further expands NLQ by repurposing the extensive narrations within Ego4D as queries, thereby enhancing model performance~\cite{groundnlq}. However, these narrations are not directly applicable to question-answering (QA) tasks. To bridge this gap, we introduce a generation pipeline that transforms these narrations into structured QA pairs. Additionally, we establish a multi-modal generative model capable of temporally localizing and answering a language query given long, egocentric videos.

\vspace{2pt} \noindent \textbf{Video question answering.}
Video question answering (VideoQA) entails generating responses to natural language queries by analyzing video content. This challenging task requires a detailed understanding of both visual and textual information. The advent of VideoQA datasets has catalyzed advancements in VideoQA research and benchmarking. For instance, ActivityNet-QA~\cite{activitynet-qa}, which focuses on a variety of human activities, facilitates the evaluation of a model's ability to interpret complex actions and interactions. In contrast, How2QA~\cite{how2qa}, derived from instructional videos, emphasizes understanding sequential processes. NextQA~\cite{nextqa} stands out by concentrating on causal and temporal reasoning in videos. These datasets typically include short videos, thereby limiting their relevance to real-world situations. In response, \qaego4d~\cite{qaego4d} offers a long-form VideoQA benchmark featuring over a thousand egocentric videos with an average length of 8.2 minutes, each annotated with open-ended answers based on the aforementioned NLQ data. Our proposed method achieves state-of-the-art performance on this benchmark.

Annotating VideoQA datasets is labor-intensive and expensive~\cite{justask}, while insufficient training data often results in over-fitting. To address this, automatic generation of VideoQA data has been investigated. For example, JustAsk~\cite{justask} generates QA pairs from transcribed speech using pre-trained language models, substantially expanding the dataset size. More recently, Large language models (LLMs) have shown remarkable proficiency in task processing and reasoning. Innovative studies like LLaVA~\cite{llava} and MiniGPT-4~\cite{minigpt4} leverage the powerful capabilities of LLMs to generate visual instruction tuning data, achieving notable success in a range of visual-language tasks.
In our study, we exploit LLMs to transform existing narrations from the Ego4D dataset into question-answer pairs with temporal windows, facilitating multimodal understanding for egocentric videos.
A concurrent work, EgoSchema~\cite{egoschema}, also exploits LLMs for constructing QA pairs. Compared to it, our approach includes both CloseQA and OpenQA, offering greater real-world applicability. Moreover, EgoSchema aims to summarize entire videos, while our method emphasizes episodic memory, focusing on recalling specific fragments for fine-grained queries.

\vspace{2pt} \noindent \textbf{Egocentric video understanding.} 
Egocentric video understanding, a rapidly evolving field, focuses on analyzing videos captured by wearable cameras. This field boosts a wide range of applications, including robotics, healthcare, augmented reality, and assistance for individuals with visual impairments. Various datasets are accessible to support research in this domain, including EPIC-KITCHENS~\cite{epic-kitchens}, which contains videos of kitchen activities; Charades-Ego~\cite{charades-ego}, featuring various everyday tasks; and Ego4D~\cite{ego4d}, which provides a global collection of diverse egocentric videos. These resources have raised emerging research problems such as human-object interaction~\cite{nagarajan2019grounded}, action recognition~\cite{kazakos2019epic}, and predictive modeling~\cite{girdhar2021anticipative}, \textit{etc}. In this work, we delve into the complex task of grounded question answering, which demands temporally localizing a segment from an untrimmed egocentric video that corresponds to a given question, and producing an answer in natural language.

\begin{figure*}[t]
\centerline{\includegraphics[width=0.9\linewidth]{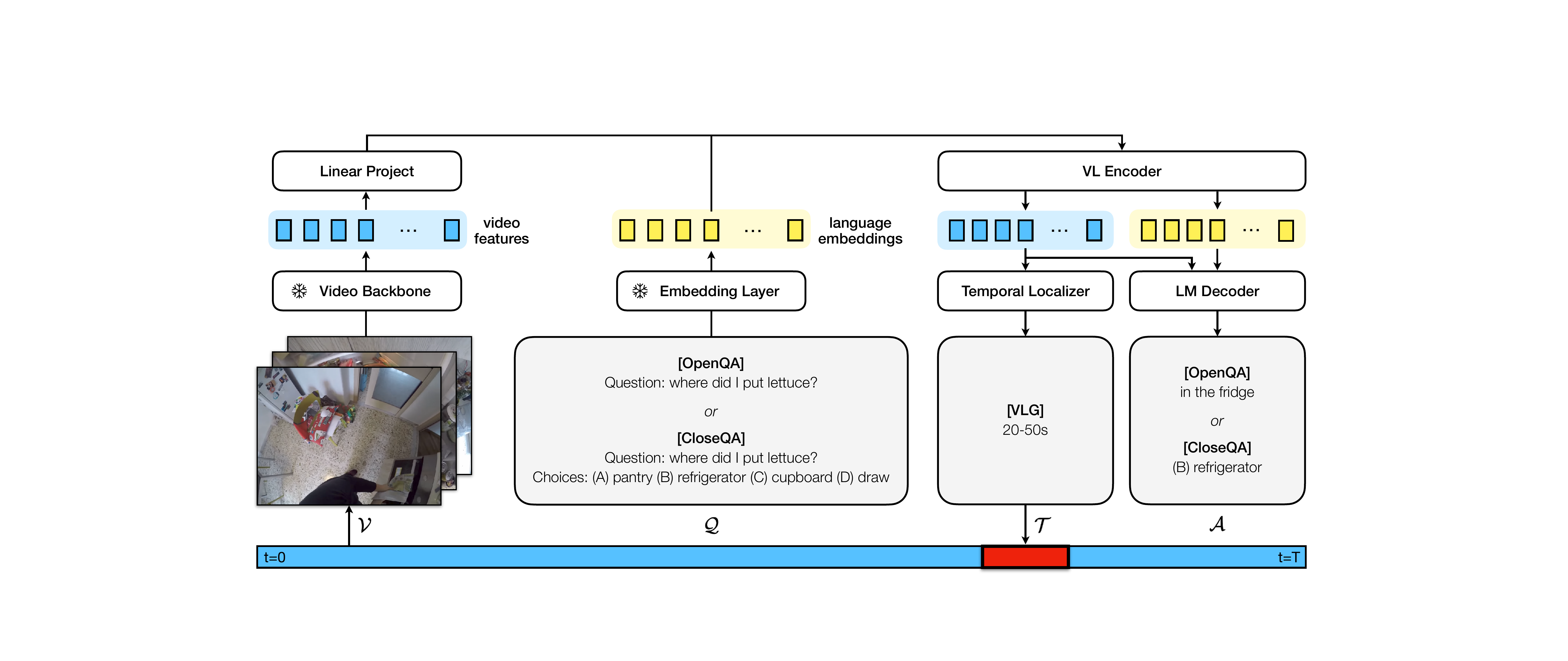}}
\caption{\textbf{Overview of \ourmethod}. It addresses three tasks: OpenQA, CloseQA, and VLG. The model processes a video $\mathcal{V}$ and a question $\mathcal{Q}$, to reason about the relevant temporal window $\mathcal{T}$ and the answer $\mathcal{A}$. Initially, a frozen video backbone encodes $\mathcal{V}$ and maps it into the language embedding space. Simultaneously, $\mathcal{Q}$ undergoes tokenization and is transformed through an embedding layer. These video and question embeddings are then fused using a visual-language encoder. Finally, a temporal localizer uses the resulting video features to predict $\mathcal{T}$, whereas a language decoder utilizes both video and question features, as provided by the VL encoder, to generate $\mathcal{A}$.} \label{fig:framework} 
\vspace{-8pt}
\end{figure*}

\section{Method} \label{sec:method}

This paper investigates the problem of grounded question answering in long egocentric videos, {\em i.e.}, the simultaneous localization and answering of questions. In Sec.~\ref{sec:method_task}, we begin by formally defining the task. In Sec.~\ref{sec:method_model}, we introduce our model, \ourmethod, that enables temporally grounding of visual questions and generates answers in either free-form language or a multi-choice format. In Sec.~\ref{sec:method_generate_qa}, we describe an automatic QA generation pipeline that leverages Large Language Models (LLMs) to transform narrations into QA pairs with temporal windows, a strategy proven to mitigate overfitting caused by limited training data in existing QA dataset on egocentric videos~\cite{qaego4d}. Lastly, in Sec.~\ref{sec:training}, we detail the multi-task training procedure for our model.

\subsection{Task Definition} \label{sec:method_task}

In general, we are interested in the task of generating open-ended answers to natural language questions, with an emphasis on the challenges of temporal grounding and contextual visual-language understanding.

Considering an egocentric video $\mathcal{V} \in \mathbb{R}^{N \times H \times W \times 3}$ and a question $\mathcal{Q} := \{ q_1, q_2, \ldots, q_M \}$ as inputs -- where 
$N$ denotes the number of frames, $H$ and $W$ are the dimensions of each frame, and $M$ is the length of the query -- our objective is to construct a model $\Phi$ that simultaneously performs question grounding and answering:
{
\setlength\abovedisplayskip{3pt}
\setlength\belowdisplayskip{5pt}
\begin{align}
[\mathcal{T}, \mathcal{A}] = \Phi(\mathcal{V}, \mathcal{Q}).    \label{equ:1}
\end{align}
}The temporal window $\mathcal{T}:=(s,~e)$, defined by its start time $s$ and end time $e$, pinpoints a specific segment of the video that is most relevant to the posed question, aligning with the concept of Video Language Grounding (\textbf{VLG}). Moreover, $\mathcal{A}$ is the generated responses, which can be in free-form language for open-ended question answering (\textbf{OpenQA}) or selected from multiple choices for close-ended question answering (\textbf{CloseQA}). Our proposal involves the concurrent training of the model on these three tasks.

\subsection{A Multi-tasking Architecture} \label{sec:method_model}

In Fig.~\ref{fig:framework}, we present the architecture of our proposed \textbf{\ourmethod}, comprising five main components: a language embedding layer, a video feature encoder, a linear projection layer, a visual-language encoder, and a dual-headed decoder for temporal localization and answer generation. This section describes each component in detail.

\vspace{3pt} \noindent\textbf{Language embedding layer.} 
This layer transforms the tokenized query into vector embeddings: $\mathcal{Q}^{\prime} = \phi_{\text{emb}} (\mathcal{Q})$. Specifically, in OpenQA, the term ``query'' refers to the questions being asked, whereas in CloseQA, a set of $K$ candidate answers is appended to the question. 

\vspace{3pt} \noindent\textbf{Video encoder and projection layer.} 
We utilize a frozen encoder, $\psi_{\text{v}}$, to extract features from the video sequence. These features are then mapped to the language embedding space by a linear projection layer: $\mathcal{V}^{\prime} = \phi_{\text{proj}} \circ \psi_{\text{v}}(\mathcal{V})$.

\vspace{3pt} \noindent\textbf{Visual-language encoder.} 
Here, we use several Transformer encoder layers~\cite{transformer} that accept the projected video features and query embeddings as inputs, and fuse the visual-language information: $[\hat{\mathcal{Q}}, \hat{\mathcal{V}}] = \psi_{\text{vl}}(\mathcal{Q}^{\prime}, \mathcal{V}^{\prime})$. 

\vspace{3pt} \noindent\textbf{Temporal question localizer.} 
The objective here is to identify a temporal window within the video, that is most informative for answering the specific question. Our localizer takes the updated video feature from the visual-language encoder, and predicts the temporal window, {\em i.e.}, $\hat{\mathcal{T}} = \psi_{\text{t}}(\hat{\mathcal{V}})$. Specifically, we adopt a similar module as in GroundNLQ~\cite{groundnlq} and ActionFormer~\cite{actionformer}, which consists of a classification head and a regression head. The classification head outputs a probability score for each timestamp's relevance to the question, while the regression head estimates the boundary distances from the current timestamp. 

\vspace{3pt} \noindent\textbf{Language decoder.} 
To generate answers to specific visual questions, we use a causal Transformer decoder. This decoder cross-attends to the output video and question features from the visual-language encoder and generates the answer in an auto-regressive manner: 
$\hat{\mathcal{A}} = \psi_{\text{d}}(\hat{\mathcal{Q}}, \hat{\mathcal{V}})$.

\subsection{Generate QA from Narrations} \label{sec:method_generate_qa}

\begin{figure*}[ht]
\centerline{\includegraphics[width=.96\linewidth]{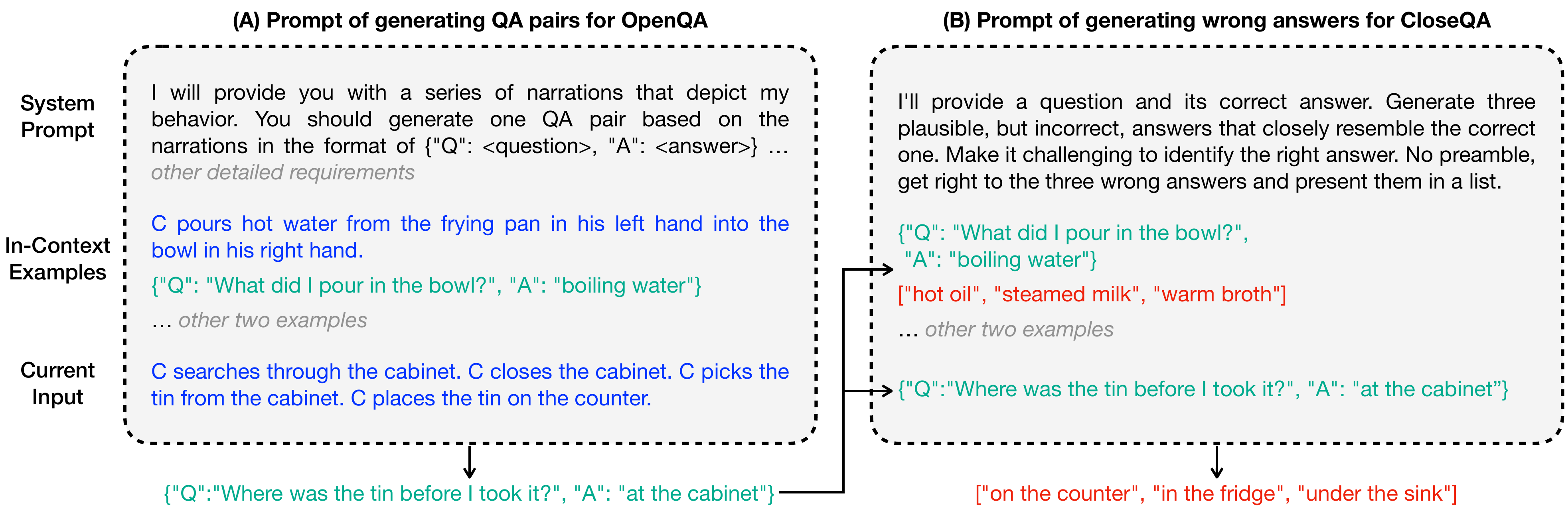}}
\vspace{-3pt}
\caption{
\textbf{The prompts for generating OpenQA and CloseQA training data with Llama2.}
(A) First, we generate question-answer pairs using consecutive narration sentences from Ego4D. (B) Next, we generate three plausible yet incorrect answers for each question-answer pair to construct data for the CloseQA task. We provide in-context examples to enhance the generation quality.
} \label{fig:prompt} 
\vspace{-7pt}
\end{figure*}

To train our model in concurrent query grounding and answering, as outlined in Equation~\ref{equ:1}, we utilize the Ego4D dataset~\cite{ego4d}. This dataset comprises a vast collection of egocentric videos, each annotated with detailed, timestamped narrations describing the activities of the person wearing the camera, with an average of 13.2 sentences per minute. Our goal is to exploit these high-quality narrations to create an automated pipeline that generates QA training samples using large language models~(LLMs). 

\vspace{3pt}
\noindent\textbf{Estimating temporal windows for narrations.} 
In an egocentric video $\mathcal{V}_i$, narrations are represented by the set $\{(\mathcal{N}_j, t_j)\}$, where $\mathcal{N}_j$ is a narration sentence and $t_j$ is its timestamp. 
To determine the temporal windows, we adopt a strategy akin to that in EgoVLP~\cite{egovlp}:
{
\setlength\abovedisplayskip{3pt}
\setlength\belowdisplayskip{3pt}
\begin{equation}
  \mathcal{T}_j = \left( t_j - \frac{\beta_i}{2\alpha}, \quad t_j + \frac{\beta_i}{2\alpha} \right),
\end{equation}
}where $\beta_i$ denotes the average interval between the timestamps of consecutive narrations, and $\alpha$ is the average of all $\beta_i$ values across videos. Essentially, these temporal windows are defined based on the dataset statistics.

\vspace{3pt}
\noindent\textbf{Generating OpenQA data.} 
We use an LLM to generate QA pairs from consecutive narration sentences. Considering that individual narration sentences are relatively short~(7.4 words on average) and may lack sufficient information for generating meaningful questions, we propose to group consecutive sentences that collectively convey a complete context. Specifically, we segment the chronologically arranged narrations of a video into chunks. These chunks are based on either up to 5 sentences or a maximum duration of 30 seconds, whichever is reached first. For each chunk, we prompt the LLM to generate one QA pair and merge the associated temporal windows, resulting in a $(\mathcal{Q}, \mathcal{A}, \mathcal{T})$ pair. As depicted in Fig.~\ref{fig:prompt}~(A), the prompt comprises the chunk's narrations, detailed instructions, and three in-context examples to enhance the generation quality. 

Utilizing the Llama2-13B model~\cite{llama2} on an NVIDIA A100 (80GB) GPU, we can generate approximately 20K QA pairs per hour, which is significantly more efficient than manual annotation. We apply this method to the training split of Ego4D$_\texttt{v2}$ Episodic Memory dataset. Consequently, we have created \textbf{\ourdataset}, a grounded QA dataset containing 5,389 egocentric videos and 303K samples, as detailed in Tab.~\ref{tab:datasets}.

\vspace{3pt}
\noindent\textbf{Generating CloseQA data.}
We prompt the LLM to generate three options that appear valid but are ultimately incorrect for a given question-answer pair. The constructed prompt is illustrated in Fig.~\ref{fig:prompt}~(B). We apply this procedure to augment \ourdataset~and \qaego4d, enabling the training and evaluation of models in a multi-choice scenario. 
The generation speed reaches 40K samples per hour.

\vspace{3pt}
\noindent\textbf{Filtering CloseQA test set.}
The LLM may generate implausible choices. To maintain the rigor of the CloseQA task, we filter out questions from the \qaego4d~test set that are easily answerable without video context. Specifically, we train a text-only ``blind'' model to identify and remove questions that are consistently answered correctly across ten trials with different seeds. 
Additionally, we perform rigorous human verification by eliminating samples that contain incorrect answers or temporal windows.
The resulting \textbf{\qaego4d$_\texttt{Close}$} serves as a more refined testing ground. This ensures that models being evaluated truly require video content analysis to answer the questions correctly, thereby emphasizing the visual aspect of CloseQA.

\subsection{Multi-task Training} 
\label{sec:training}

Our model is designed to simultaneously address three tasks: open-ended question answering (OpenQA), close-ended question answering (CloseQA), and video-language grounding (VLG).

\vspace{3pt} \noindent \textbf{Training for question-answering.}
Training alternates between OpenQA and CloseQA to ensure proficiency in both question-answering formats. For OpenQA, inputs follow the format \texttt{question:~<question>? video: <video feature>}. While for CloseQA, inputs are structured as \texttt{question: <question>? choices: <choices>. video: <video feature>}. To avoid memorization of answer positions, candidate answers are randomly shuffled. Moreover, the model is tasked to not only identify the correct option but also generate the associated answer, increasing training complexity. Cross-entropy loss is employed for both tasks, expressed as $\mathcal{L}_\text{QA} = \mathcal{L}_\text{ce}(\mathcal{A}, \hat{\mathcal{A}})$.

\vspace{3pt} \noindent \textbf{Training for video-language grounding.}
Concurrent with question-answering tasks, our model undergoes training on the VLG task. We employ temporal jittering~\cite{naq} to augment temporal windows through random scaling and shifting. The loss function is a combination of binary Focal loss~\cite{focal} and DIoU loss~\cite{diou}, formulated as $\mathcal{L}_\text{VLG} = \mathcal{L}_\text{focal}( \mathcal{T},~\hat{\mathcal{T}}) + \mathcal{L}_\text{DIoU}(\mathcal{T},~\hat{\mathcal{T}})$.

The final loss is a weighted sum: $\mathcal{L} = 0.5 \times \mathcal{L}_\text{VLG} + 0.5 \times \mathcal{L}_\text{QA}$. Incorporating the VLG task into training enhances the visual-language encoder's capability to distill relevant information from videos, thereby boosting QA performance. Our model can be exclusively trained on VLG by freezing the LM decoder, or on a QA task by freezing the temporal localizer. Relevant experiments are provided in Sec.~\ref{sec:experiments_ablations}.
\section{Experiments}
\label{sec:experiments}

\begin{table}[t]
\setlength{\tabcolsep}{1.0mm}
\centering
\resizebox{\columnwidth}{!}{
\begin{tabular}{llcccccc}
    \toprule
     & \multirow{2}{*}{Dataset} & \multirow{2}{*}{\# Video} & \multirow{2}{*}{\# Sample} & & \multicolumn{3}{c}{Supported Task} \\
     \cmidrule{6-8}
     & & & & & OpenQA & CloseQA & VLG \\
    \midrule
    \parbox[t]{2mm}{\multirow{2}{*}{\small\rotatebox[origin=c]{90}{train}}}
     & \qaego4d & 997 & 11K & & $\checkmark$ & $\checkmark$ & $\checkmark$ \\
     & \textbf{\ourdataset} & 5,389 & 303K & & $\checkmark$ & $\checkmark$ & $\checkmark$ \\
    \hdashline
    \parbox[t]{2mm}{\multirow{2}{*}{\small\rotatebox[origin=c]{90}{val}}}
     & \qaego4d & 162 & 1913 & & $\checkmark$ & -- & -- \\
     & NLQ$_\texttt{v2}$ & 415 & 4,552 & & -- & -- & $\checkmark$ \\
    \hdashline
    \parbox[t]{2mm}{\multirow{3}{*}{\small\rotatebox[origin=c]{90}{test}}}
     & \qaego4d & 166 & 1,850 & & $\checkmark$ & -- & -- \\
     & \textbf{\qaego4d$_\texttt{Close}$} & 148 & 500 & & -- & $\checkmark$ & -- \\
     & NLQ$_\texttt{v2}$ & 333 & 4,004 & & -- & -- & $\checkmark$ \\
    \bottomrule
\end{tabular}}
\vspace{-3pt}
\caption{
\textbf{A summary of detailed dataset statistics.} 
Both the \qaego4d~and our \ourdataset~datasets support training on OpenQA, CloseQA, and VLG tasks. Hyper-parameters are picked based on the validating results on \qaego4d~and NLQ$_\texttt{v2}$, while models' performance is evaluated on corresponding test sets.}
\label{tab:datasets}
\vspace{-5pt}
\end{table}

\subsection{Dataset and Metrics}

\noindent\textbf{Natural Language Query~(NLQ)}~\cite{ego4d} is a prominent example of the video language grounding task. 
The second version of this benchmark, NLQ$_\texttt{v2}$, comprises 1,659 video clips, paired with 17.3K natural language queries and corresponding temporal windows.  
It is split into train, validation, and test sets, containing 11.3K, 3.9K, and 4K pairs, respectively. 
For evaluation, we use Recall@k, IoU=m, 
where $k \in \{1, 5\}$ and $m \in \{0.3, 0.5\}$. 
The primary metric for the NLQ challenge is \textbf{Mean Recall@1}, computed as the average of Recall@1, IoU=0.3 and Recall@1, IoU=0.5.

\vspace{3pt}
\noindent\textbf{NaQ}~\cite{naq} augments NLQ by repurposing the extensive narrations with Ego4D as queries, including 5,389 video clips and 945K training samples.

\vspace{3pt}
\noindent\textbf{\ourdataset} is our contributed pre-training dataset, containing the same video clips as NaQ, while featuring 303K question-answer pairs with temporal windows.

\vspace{3pt}
\noindent\textbf{\qaego4d~\cite{qaego4d}}
expands the NLQ benchmark by manually annotating open-ended answers on its train and validation sets. 
It consists of 1,325 video clips and 14.5K data samples, further divided into 10,746 training, 1,913 validation, and 1,850 testing samples.
It adopts Accuracy and machine translation metrics including ROUGE-L (f-score)~\cite{rouge}, METEOR~\cite{meteor}, and BLEU-4~\cite{bleu}. 
In our experiments, we exclude BLEU-4 because the majority (around  $80\%$) of answers in \qaego4d are under three words in length, and Accuracy as it's not effective due to language ambiguity in open-ended answers. 
To further address such ambiguity, we choose sentence similarity (\textbf{Sim.})~\cite{sentencebert} as the primary metric, 
which maps sentences to a learned embedding space to calculate cosine similarity. Specifically, we utilize the Sentence Transformers library and the \texttt{all-MiniLM-L6-v2} language model to perform the mapping.

\vspace{3pt}
\noindent\textbf{\qaego4d$_\texttt{Close}$.}
As detailed in Sec.~\ref{sec:method_generate_qa}, we have augmented \qaego4d with a close-ended question answering (CloseQA) testing set. We run a model five times on this set with different seeds and calculate the \textbf{Accuracy} metric.

\subsection{Implementation Details} \label{sec:implementation}

\noindent\textbf{Video backbone features.} 
Recent studies, particularly InternVideo~\cite{internvideo-ego4d} and GroudNLQ~\cite{groundnlq}, have utilized features from multiple video backbones to improve performance. To ensure a fair comparison, we use identical video features: EgoVLP, InternVideo-text, and InternVideo-verb. 
We concatenated these features along the channel dimension, forming 2304-dimensional feature vectors for each time step. Unless otherwise specified, we uniformly sample 1,200 vectors from these features as model input, which corresponds to an average of 8.2 minutes of video clips.

\vspace{3pt}
\noindent\textbf{Model configurations.} 
We use an instruction-tuned version of Flan-T5~\cite{flan-t5,t5} as the language model. Our experiments involve its two variants: we conduct ablation studies using \texttt{Flan-T5-Small} (denoted as \ourmethod$_\texttt{S}$) and make final comparisons using \texttt{Flan-T5-Base} (denoted as \ourmethod$_\texttt{B}$). 
Our temporal localizer is adapted from ActionFormer~\cite{actionformer}, 
without using multi-scale pyramid features. 
This localizer comprises a classification head and a regression head, 
each has two layers of 1D convolution with layer normalization and ReLU activation in between. 

\vspace{3pt}
\noindent\textbf{Training details.} 
We train all models with the AdamW optimizer~\cite{adamw}, setting $\beta_1=0.9, \beta_2=0.999$, a learning rate of $1 \times 10^{-4}$, 
and no weight decay.
The language embedding layer of Flan-T5 is fixed during training. 
Experiments are carried out on 4 NVIDIA A100 (80GB) GPUs, 
with gradient accumulation to maintain a consistent global batch size of 128. The training process is limited to 100 epochs, with early stopping based on the validation performance.

\subsection{QA Baselines} \label{sec:experiments_baselines}

As baseline models, we adopt the same models used in~\cite{qaego4d} and 
introduce several improvements for a fair comparison.

\vspace{3pt}
\noindent\textbf{BlindVQA} fine-tunes a \texttt{T5-Base} language model
to answer questions without using video input. 
Essentially, BlindVQA serves as a language-only model to understand whether visual signals are essential to a specific question.

\vspace{3pt}
\noindent\textbf{SimpleVQA} enhances BlindVQA by incorporating visual capabilities. Here, video features are mapped to the language space and concatenated with question embeddings from an LM encoder. An LM decoder then generates answers given the merged features. Our proposed \ourmethod~model differs from SimpleVQA, 
by conducting visual-language fusion in the encoder and adopting VLG supervision on the fused video features.

\vspace{3pt}
\noindent\textbf{SimpleVQA+} builds on SimpleVQA by adding a ranking loss on LM Decoder's cross attention. Like our approach, SimpleVQA+ uses VLG supervision to emphasize the model's attention on question-relevant video segments. However, it cannot predict temporal windows, hindering the assessment of its grounding ability. Additionally, its performance falls short compared to our \ourmethod.

\vspace{3pt}
\noindent\textbf{Rehearsal Memory (RM)}~\cite{rehearsal} compresses long videos into a fixed-size memory.
It segments a lengthy video into uniform parts, each processed by a Transformer encoder. Then, a recurrent module sequentially attends each segment feature to update the memory state. 
RM pretrains the memory state using reconstruction as a proxy task and further fine-tunes on the QA task.

\vspace{3pt}
\noindent\textbf{Improved baselines.} 
To ensure a fair comparison, we make several enhancements to the baseline models: (i) Replacing the original SlowFast~\cite{slowfast} features with EgoVLP and InternVideo features, as specified in Sec.~\ref{sec:implementation}; (ii) Upgrading T5 to Flan-T5 and freezing word embeddings during training; (iii) Increasing the batch size to 128 and adjusting the learning rate to $1 \times 10^{-4}$. These modifications have consistently boosted the baseline model performance.

\begin{table*}[t]
\centering
\footnotesize
\resizebox{0.95\linewidth}{!}{
\begin{tabu}{cl c cccc ccc cc}
    \toprule
    \multirow{2}{*}{} & \multirow{2}{*}{Model} & Additional Data &\phantom{a}& \multicolumn{2}{c}{Additional Task} &\phantom{a}& \multicolumn{3}{c}{OpenQA} &\phantom{a}& CloseQA \\
     \cmidrule{3-3} \cmidrule{5-6} \cmidrule{8-10} \cmidrule{12-12}
     & & \ourdataset & & CloseQA & VLG && \textbf{Sim.} & ROUGE & METEOR && \textbf{Accuracy} \\
    \midrule
    (A) & \textbf{\ourmethod$_\texttt{S}$} & -- & & -- & -- && 54.9 & 27.9 & 18.8 && -- \\
    (B) & \textbf{\ourmethod$_\texttt{S}$} & -- & & $\checkmark$  & -- && 54.8 & 27.7 & 18.7 && 39.5$\pm$0.5 \\
    (C) & \textbf{\ourmethod$_\texttt{S}$} & -- & & $\checkmark$  & $\checkmark$  && 55.6 & 29.0 & 19.8 && 40.8$\pm$1.0 \\
    (D) & \textbf{\ourmethod$_\texttt{S}$} & $\checkmark$ & & $\checkmark$  & -- && 56.1 & 28.8 & 20.1 && 47.2$\pm$0.5 \\
    (E) & \textbf{\ourmethod$_\texttt{S}$} & $\checkmark$ & & $\checkmark$  & $\checkmark$  && 57.7 & 30.2 & 21.2 && 48.7$\pm$0.4 \\
    \rowfont{\color{gray}}(F) & Oracle        & $\checkmark$ & & $\checkmark$ & -- && 58.4 & 30.9 & 21.9 && 53.5$\pm$0.7\\
    \hdashline
    (G) & SimpleVQA$_\texttt{S}$  & -- & & $\checkmark$  & -- && 54.9 & 28.0 & 19.0 && 41.3$\pm$0.4 \\
    (H) & SimpleVQA$_\texttt{S}$  & $\checkmark$ & & $\checkmark$  & -- && 56.1 & 28.8 & 20.2 && 47.1$\pm$0.3 \\
    \hdashline
    (I) & SimpleVQA+$_\texttt{S}$ & -- & & $\checkmark$  & \scalebox{0.75}{$\bigstar$} && 54.7 & 27.9 & 19.0 && 39.3$\pm$0.6 \\
    (J) & SimpleVQA+$_\texttt{S}$ & $\checkmark$ & & $\checkmark$  & \scalebox{0.75}{$\bigstar$} && 55.4 & 28.1 & 19.5 && 42.0$\pm$0.7 \\
    \bottomrule
\end{tabu}}
\vspace{-5pt}
\caption{\textbf{Ablation study on \qaego4d and \qaego4d$_\texttt{Close}$ test sets.}
``Additional Data'': adding training data beyond \qaego4d.
``Additional Task'': incorporating training tasks beyond OpenQA.
``Sim.'': the Sentence Similarity metric. 
``\textcolor{gray}{Oracle}'' represents a variant of \ourmethod$_\texttt{S}$, taking only question-relevant video segments as input to bypass the need for temporal grounding, thereby establishing the upper-bound performance.
SimpleVQA+ leverages VLG supervision but cannot solve the VLG task, indicated by ``\scalebox{0.8}{$\bigstar$}''.
}
\label{tab:ablation_qa}
\vspace{-10pt}
\end{table*}
\begin{table}[t]
\setlength{\tabcolsep}{1.0mm}
\centering
\resizebox{\linewidth}{!}{
\begin{tabu}{lc ccccc c}
\toprule
   \multirow{2}{*}{Training} & \multirow{2}{*}{\ourdataset} &$\ $& \multicolumn{3}{c}{OpenQA} && CloseQA \\
    \cmidrule{4-6} \cmidrule{8-8}
                &&& \textbf{Sim.} & ROUGE & METEOR && \textbf{Accuracy}  \\ 
\midrule
Two-stage         & -- && 54.7 & 27.3 & 18.4 && 39.3$\pm$0.8 \\
\textbf{Unified}  & -- && 55.6 & 29.0 & 19.8 && 40.8$\pm$1.0 \\
\hdashline
Two-stage         & $\checkmark$ && 56.0 & 28.3 & 19.9 && 46.4$\pm$0.7 \\
\textbf{Unified}  & $\checkmark$ && 57.7 & 30.2 & 21.2 && 48.7$\pm$0.4 \\
\bottomrule
\end{tabu}}
\vspace{-5pt}
\caption{
\textbf{Effect of the unified training method.}
The "Two-stage" method separately trains two \ourmethod$_\texttt{S}$ models: one for the VLG task, and the other for QA tasks using relevant video segments. During inference, it uses the grounding results from the first model in the question-answering process of the second model.
}
\label{table:ablation_unified}
\end{table}
\begin{table}[t]
\setlength{\tabcolsep}{1.5mm}
\centering
\resizebox{\linewidth}{!}{
\begin{tabu}{ccc|cc}
    \toprule
    \qaego4d & \ourdataset & NLQ$_\texttt{v2}$+NaQ & \textbf{Mean R@1} & Mean R@5 \\
    \midrule
    $\checkmark$ & -- & -- & 8.8 & 20.0 \\
    $\checkmark$ & $\checkmark$ & -- & 18.4 & 37.2 \\
    $\checkmark$ & $\checkmark$ & $\checkmark$ & 20.9 & 42.5 \\
    \bottomrule
\end{tabu}}
\vspace{-5pt}
\caption{
  \textbf{Data scaling effect on the NLQ$_\text{v2}$ val set.}
  We train our \ourmethod$_\texttt{S}$ model on OpenQA, CloseQA, and VLG tasks with different training data, and evaluate its VLG performance.
}
\label{tab:ablation_vlg}
\vspace{-0.5cm}
\end{table}

\subsection{Ablations} \label{sec:experiments_ablations}

In this section, we conduct experiments to investigate the effect of our proposal,
for example, joint training of multiple tasks, integrating \ourdataset, etc.

\vspace{3pt}
\noindent\textbf{Integrating the CloseQA task.} 
As presented in Tab.~\ref{tab:ablation_qa}~(A-B), 
simultaneously training OpenQA and CloseQA tasks, 
despite their varying input-output formats, marginally impacts OpenQA performance. However, this integration offers a more comprehensive and reasonable method for assessing the model's question-answering capabilities. 
Thus, we integrate CloseQA in training as default.

\vspace{3pt}
\noindent\textbf{Integrating the VLG task.} 
As shown in Tab.~\ref{tab:ablation_qa}, incorporating VLG task indeed improves question-answering performance, {\em e.g.}, \ourmethod$_\texttt{S}$'s Sentence Similarity increases from 54.8 to 55.6 when trained on \qaego4d~(B-C) and increases from 56.1 to 57.7 when trained on both \qaego4d~and \ourdataset~(D-E),
demonstrating the effectiveness of our proposed multi-task training approach.

Conversely, SimpleVQA+$_\texttt{S}$ utilizes VLG supervision to direct the LM Decoder's cross attention towards question-related video segments. However, this approach results in diminished QA performance (G to I and H to J). This suggests that the complexity of the VLG task exceeds the capacity of the cross-attentions.

\vspace{3pt}
\noindent\textbf{Unified v.s. separate training.} An alternative to our unified model is training two separate models, one for temporal grounding and the other for question-answering on the grounded video clip. Results in Tab.~\ref{table:ablation_unified} validate the effectiveness of our unified training method.

\begin{figure}[t]
\centerline{\includegraphics[width=\columnwidth]{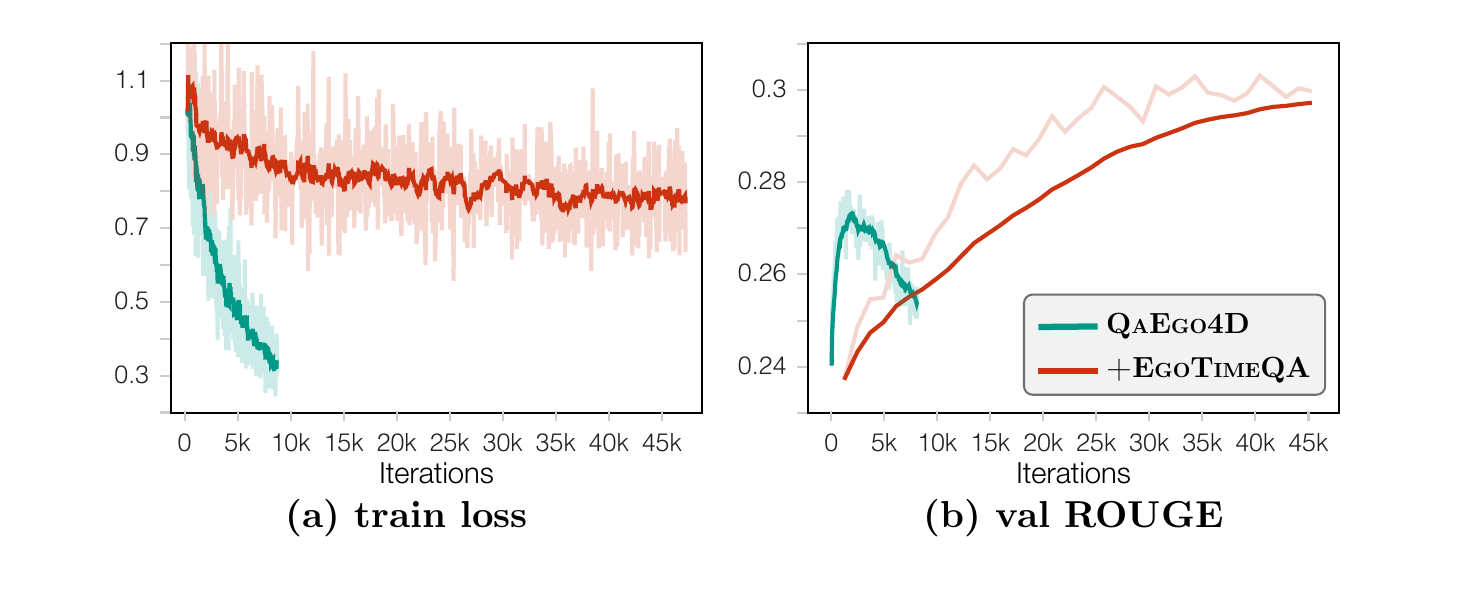}}
\vspace{-5pt}
\caption{
\textbf{Training and validation curves of \ourmethod$_\texttt{S}$.}
The limited training data of \qaego4d results in severe overfitting, which is effectively mitigated by our generated \ourdataset.
} \label{fig:overfitting} 
\vspace{-0.5cm}
\end{figure}

\vspace{3pt}
\noindent\textbf{Incorporating \ourdataset~data.} Our data generation method produces 303K samples, a 30-fold increase over the \qaego4d~training set, resulting in notable performance gains in QA and VLG tasks. The QA metrics for \ourmethod$_\texttt{S}$ demonstrate significant enhancements, as evidenced in Tab.~\ref{tab:ablation_qa} (B to D and C to E). Similar improvements are observed for SimpleVQA$_S$ (G-H) and SimpleVQA+$_S$ (I-J), confirming the generality and effectiveness for \ourdataset. In Tab.~\ref{tab:ablation_vlg}, \ourdataset~also boosts VLG recall by a large margin, which is further amplified with NLQ and NaQ data. Notably, as depicted in Fig.~\ref{fig:overfitting}, the value of \ourdataset~is even more evident in overcoming overfitting.

\vspace{3pt}
Combining the above enhancements (A-E in Tab.~\ref{tab:ablation_qa}), our method closely approaches the oracle upper bound (F), with the main gap due to imperfect temporal grounding.

\subsection{Comparison with State-of-the-art}

\begin{figure*}[t]
\centerline{\includegraphics[width=\linewidth]{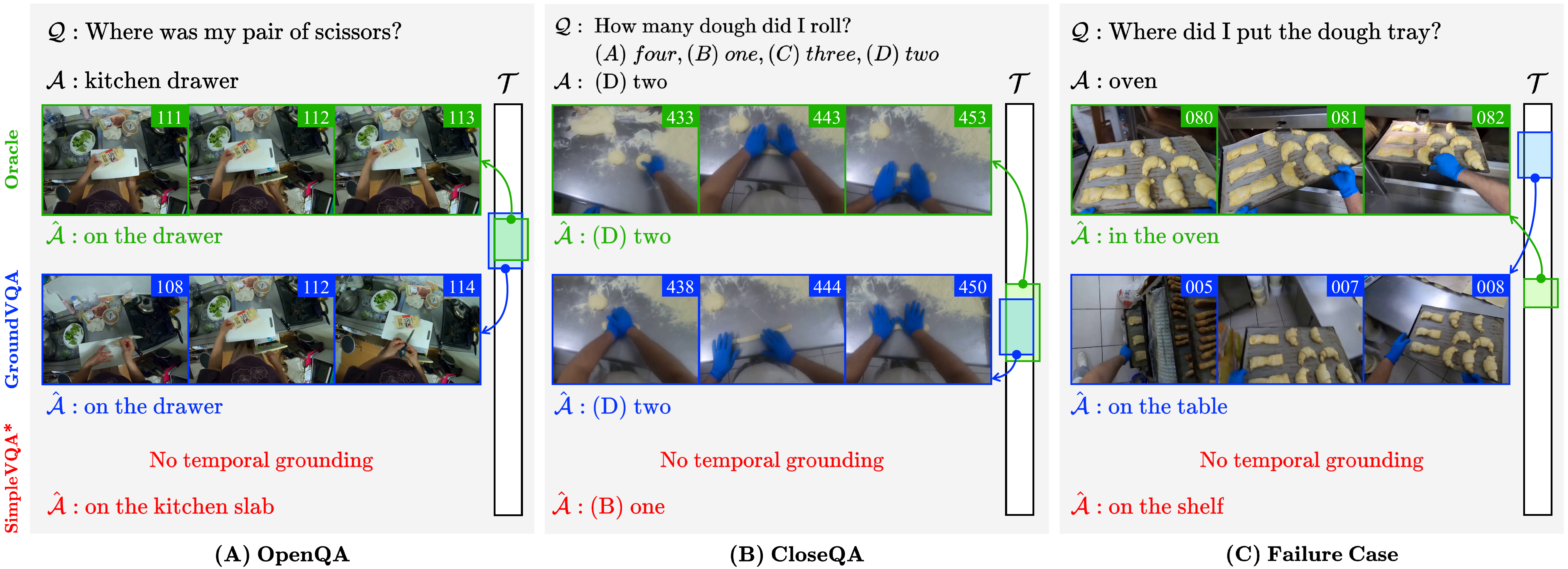}}
\vspace{-5pt}
\caption{
\textbf{Qualitative examples.}
In our demonstration, we compare three models: the \textcolor{mygreen}{Oracle} baseline, our \textcolor{blue}{\ourmethod}, and \textcolor{red}{SimpleVQA$^*$}. Each column presents a sample that includes the query $\mathcal{Q}$, the ground truth answer $\mathcal{A}$, three frames from the grounded video segment, and the predicted answer $\hat{\mathcal{A}}$. Additionally, each column illustrates the video's time span and the predicted temporal window $\mathcal{T}$, with \textcolor{mygreen}{Oracle}'s temporal window serving as the ground truth. 
Note that SimpleVQA$^*$ is incapable of predicting the temporal window.
}
\label{fig:visualization} 
\end{figure*}

In this section, we compare our model to the state-of-the-art on OpenQA, CloseQA, and VLG tasks, and present qualitative examples in Fig.~\ref{fig:visualization}.

\vspace{3pt}
\noindent\textbf{On \qaego4d.} We report results on the \qaego4d~test set in Tab.~\ref{table:qaego4d_sota}. To ensure fairness, we reproduce the other methods using identical settings (detailed in Sec.~\ref{sec:experiments_baselines}). 
Our model achieves the best performance, outperforming prior works by a large margin. 

\vspace{3pt}
\noindent\textbf{On NLQ$_\texttt{v2}$.} 
We then assess VLG performance on the NLQ$_\texttt{v2}$ test set.
As seen in Tab.~\ref{table:nlq_sota}, our model, without complex design or multi-scale feature pyramids~\cite{groundnlq}, matches the SOTA performance. \ourmethod$_\texttt{B}^{\dagger}$ exhibits further improvements by pre-training on NLQ$_\texttt{v2}$ and NaQ, and fine-tuning exclusively on NLQ$_\texttt{v2}$.

\begin{table}[t]
\setlength{\tabcolsep}{1.0mm}
\centering
\footnotesize
\resizebox{\linewidth}{!}{
\begin{tabu}{l ccc c c cc}
\toprule
   \multirow{2}{*}{Method} & \multicolumn{3}{c}{OpenQA} && CloseQA && \multirow{2}{*}{Param} \\
    \cmidrule{2-4} \cmidrule{6-6}
   & \textbf{Sim.} & ROUGE & METEOR && \textbf{Accuracy} &  \\ 
\midrule
\rowfont{\color{gray}}BlindVQA      & - & 25.9 & 17.4 && - && 247 \\
BlindVQA$^*$                             & 53.8 & 27.5 & 18.4 && 36.3$\pm$0.5 && 247 \\
\rowfont{\color{gray}}SimpleVQA     & - & 26.1 & 17.4 && - && 249 \\
SimpleVQA$^*$                            & 55.7 & 28.6 & 19.3 && 41.1$\pm$0.5 && 249 \\
\rowfont{\color{gray}}SimpleVQA+    & - & 27.1 & 18.3 && - && 249 \\
SimpleVQA+$^*$                           & 55.7 & 28.8 & 19.5 && 41.4$\pm$0.3  && 249 \\
\rowfont{\color{gray}}RM            & - & 26.6 & 17.7 && - && 368 \\
RM$^*$                                   & 54.1 & 27.3 & 18.5 && 39.9$\pm$0.8  && 368 \\
\rowcolor[RGB]{225,225,225}\textbf{\ourmethod$_\texttt{B}$}  & \textbf{58.2} & \textbf{30.4} & \textbf{21.5} && \textbf{50.2$\pm$0.5} && 252  \\
\bottomrule
\end{tabu}}
\vspace{-5pt}
\caption{
\textbf{Comparison with the state of the art on \qaego4d and \qaego4d$_\texttt{Close}$ test sets.} 
``Param'': number of parameters in millions.
\textcolor{gray}{Gray results} are reported in~\cite{qaego4d} while ``$*$'' denotes our reproducing performance with several enhancements. ``BlindVQA'' represents the lower-bound baseline, learning only language bias.
}
\label{table:qaego4d_sota}
\vspace{-0.5cm}
\end{table}

\vspace{3pt}
\noindent\textbf{Qualitative analysis.} 
Fig.~\ref{fig:visualization}(A) shows an OpenQA example. 
Our model successfully predicts the temporal window and the answer, 
while SimpleVQA$^*$ fails. 
Although our predicted answer slightly differs from the ground truth, it's still valid, highlighting the challenge of paraphrasing in evaluating open-ended answers, thus reflecting the advantage of our CloseQA task.
Fig.~\ref{fig:visualization}(B) demonstrates a CloseQA example. 
Our model shows competence in predicting a close temporal window and identifying the correct answer. On the contrary, SimpleVQA$^*$ chooses an incorrect answer, while the absence of temporal localization hinders understanding of its error source.
Fig.~\ref{fig:visualization}(C) is a failure case of our model and SimpleVQA$^*$. Yet, our model's temporal window prediction is relevant to the query, and the predicted answer is coherent with the grounded content. 
This highlights an issue of the \qaego4d and NLQ annotations,
where multiple relevant video segments and plausible answers exist, 
but only one annotation is available per query. 

\begin{table}[t]
\setlength{\tabcolsep}{1.0mm}
\renewcommand{\arraystretch}{1.1}
\centering
\footnotesize
\resizebox{\linewidth}{!}{
\begin{tabular}{l c ccc c cc}
\toprule
    \multirow{2}{*}{Method} & & \multicolumn{3}{c}{Recall@1} & & \multicolumn{2}{c}{Recall@5} \\
    \cmidrule{3-5} \cmidrule{7-8}
     &&  \textbf{Mean} & IoU=0.3 & IoU=0.5 && IoU=0.3 & IoU=0.5 \\
    \midrule
    VSLNet~\cite{vslnet} && 4.08 & 5.42 & 2.75 && 8.79 & 5.07 \\
    EgoVLP~\cite{egovlp} && 8.35 & 10.46 & 6.24 && 16.76 & 11.29 \\
    ReLER~\cite{reler} && 10.51 & 12.89 & 8.14 && 15.41 & 9.94 \\
    NaQ++~\cite{naq} && 17.67 & 21.70 & 13.64 && 25.12 & 16.33 \\
    GroundNLQ~\cite{groundnlq} && 20.91 & 24.50 & 17.31 && \textbf{40.46} & \textbf{29.17} \\
    \rowcolor[RGB]{225,225,225}\textbf{\ourmethod$_\texttt{B}$} && 19.31 & 23.65 & 14.96 && 36.19 & 24.58 \\
    \rowcolor[RGB]{225,225,225}\textbf{\ourmethod$_\texttt{B}^{\dagger}$} && \textbf{22.15} & \textbf{26.67} & \textbf{17.63} && 39.94 & 27.70 \\
    \bottomrule
\end{tabular}}
\vspace{-5pt}
\caption{
    \textbf{Comparison with the state of the art on the NLQ$_\text{v2}$ test set.}
    ``\ourmethod$_\texttt{B}$''~is simultaneously trained on all three tasks with \qaego4d~and \ourdataset~data. ``\ourmethod$^{\dagger}_\texttt{B}$'' follows NaQ++ and GroundNLQ, pre-trained solely on the VLG task with NLQ$_\texttt{v2}$ and NaQ data, and further fine-tuned on NLQ$_\texttt{v2}$.
}
\label{table:nlq_sota}
\end{table}


We present additional results in the supplementary material, including the impact of using different LLMs to generate QA data, 
a more in-depth statistical analysis of \ourdataset, additional qualitative findings, prompts for generating QA data, limitations, and future work.
\section{Conclusion}

In conclusion, this paper tackles the challenge of grounded question answering in long egocentric videos. We demonstrate the crucial role of precise temporal grounding in effective question-answering and propose a novel, unified model that concurrently tackles both tasks. To counter the risk of overfitting due to limited training data, we introduce an automated pipeline for generating extensive question-answer pairs from narrations using LLMs. Additionally, to address the challenge of evaluating open-ended answers, we present the CloseQA benchmark, ensuring more reliable evaluations. Extensive ablation studies confirm the effectiveness of our approach, which achieves state-of-the-art performance on the \qaego4d and the Ego4D-NLQ benchmarks, marking a significant advancement in the field of egocentric video understanding.

\appendix
\appendix
\onecolumn

{
\centering
\Large
\textbf{Grounded Question-Answering in Long Egocentric Videos}\\
\vspace{0.75em}
Supplementary Material\\
\vspace{2.0em}
}

\section{Llama2 vs.~ChatGPT on Data Generation}

In the main paper, we default to using \texttt{Llama2-13B-chat} for generating QA data. Here, we experiment with \texttt{ChatGPT-3.5-turbo}\footnote{Utilizing the OpenAI API: https://platform.openai.com}. To expedite the data generation and model training process, we reduce the amount of data relative to \ourdataset. Specifically, we use both LLMs to generate QA data from the NLQ$_\texttt{v2}$ training set, which includes 1.3K video clips and 221K narration sentences. Full prompts are detailed in Sec.~\ref{sec:supp_prompt}, where minor differences exist between ChatGPT's and Llams2's prompts. Consequently, Llama2 produces 92K and ChatGPT 97K data pairs.\footnote{The variation in numbers stems from differing rates of generation errors, {\em i.e.}, the generated string cannot be converted into a dictionary containing ``Q'' and ``A'' as the in-context examples.} Compared to \qaego4d, the video clips are almost identical, but the QA pairs are denser in time.
We then train \ourmethod$_\texttt{S}$ on each dataset to assess generation quality, referencing OpenQA and VLG performance. Note that for CloseQA, both training and testing data are generated by the LLMs. Thus, evaluating CloseQA on a test set produced by one LLM, such as Llama-2, would be unfair for ChatGPT because of the bias in the generation process. Therefore, we exclude CloseQA evaluation from this experiment.

As Tab.~\ref{tab:ablation_llm}~(B-C) indicates,
the model trained on data generated by \texttt{Llama2-13B-chat} slightly outperforms the one trained on \texttt{ChatGPT-3.5-turbo} data. That is to say, Llama2's capacity to generate QA pairs from narrations is comparable to, if not better than, ChatGPT.  Additionally, from a cost perspective, Llama2 is more accessible for academic research labs or companies with certain computing resources compared to ChatGPT. In terms of data scaling, the data produced by both LLMs improves the model's performance in OpenQA and VLG (from A to B and A to C). Adding more data continues to enhance performance (from C to D).
\vspace{7em}

\begin{table}[htp]
 \setlength{\tabcolsep}{1.5mm}
\renewcommand\arraystretch{1.1}
\centering
\resizebox{\linewidth}{!}{
\begin{tabular}{c lccc cccc ccc}
\toprule
&\multicolumn{4}{c}{Training Data} && \multicolumn{3}{c}{OpenQA} && \multicolumn{2}{c}{VLG}\\ 
\cmidrule{2-5} \cmidrule{7-9} \cmidrule{11-12}
&Source & \# Clip & \# Sample & Cost && \textbf{Sim.} & ROUGE & METEOR && \textbf{Mean R@1} & Mean R@5  \\
\midrule                 
(A) & \qaego4d & 1.0K & 11K & -- && 55.6 & 29.0 & 19.8 && 8.8 & 20.0 \\
(B) & + ChatGPT QA (NLQ$_\texttt{v2}$) & 1.3K & 107K & \$50 && 56.9 & 29.2 & 19.8 && 15.7 & 33.7 \\
(C) & + Llama2 QA  (NLQ$_\texttt{v2}$) & 1.3K & 103K & 5 Gh && 57.1 & 29.4 & 20.3 && 16.3 & 34.3 \\
(D) & + Llama2 QA  (EM$_\texttt{v2}$)  & 5.5K & 314K & 16 Gh && 57.7 & 30.2 & 21.2 && 18.4 & 37.2 \\
\bottomrule
\end{tabular}}
\caption{
\textbf{Effect of data scaling and using different LLMs for data generation.} We train \ourmethod$_\texttt{S}$ with different training data and report their OpenQA and VLG performance.
``ChatGPT'' denotes the \texttt{ChatGPT-3.5-turbo} model in OpenAI's API. ``Llama2'' is short for \texttt{Llama2-13B-chat}. Row D is our \ourdataset. In the ``Cost'' column, we estimate the money or time spent on generating the corresponding data, where ``Gh'' stands for GPU hours tested on NVIDIA A100 (80GB) GPUs.
}\label{tab:ablation_llm}
\end{table}

\clearpage
\section{Additional \ourdataset~Statistics}

\begin{figure}[ht]
\captionsetup{font=normalsize}
\captionsetup[sub]{font=normalsize}
\setlength{\belowcaptionskip}{0.5\baselineskip}
\centering
    \begin{subfigure}{.42\linewidth}
        \centering
        \includegraphics[width=1\linewidth]{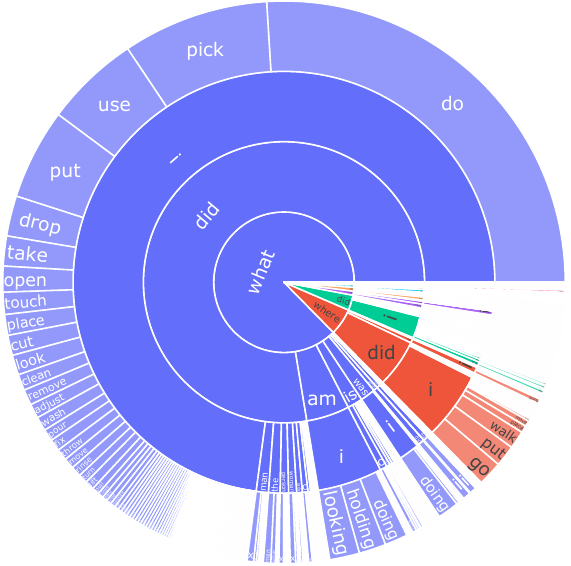}
        \caption{\textbf{Question sunburst chart.} The word order starts at the center and extends outward. A larger area indicates a higher frequency of occurrence.}
        \label{fig:supp_question_sunburst}
    \end{subfigure}%
    \hfill
    \begin{subfigure}{.56\linewidth}
        \centering
        \begin{subfigure}{\linewidth}
            \centering
            \includegraphics[width=1\linewidth]{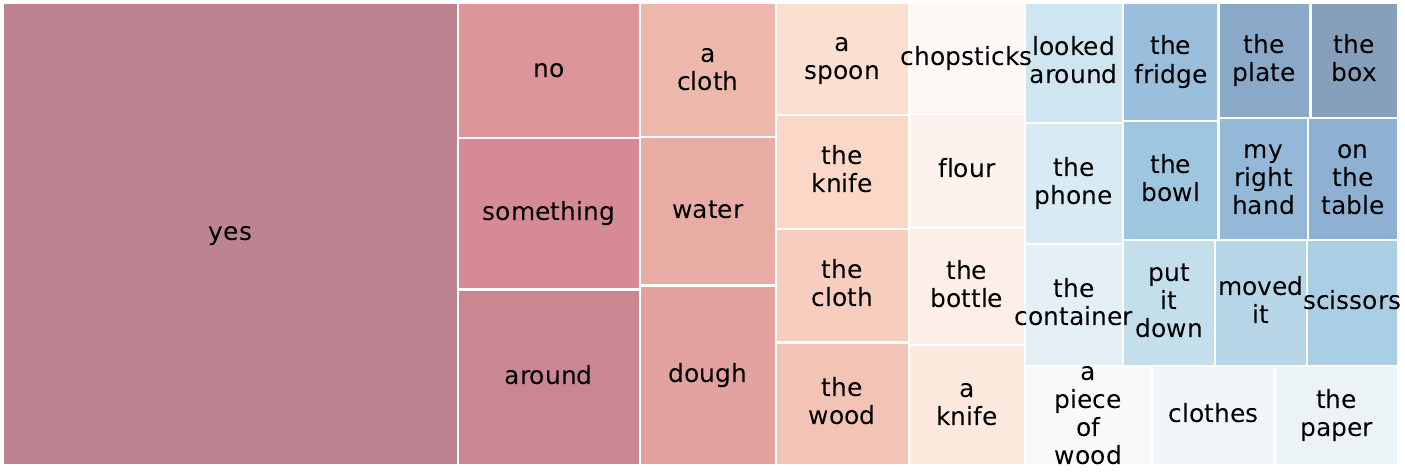}
            \caption{\textbf{Correct answer treemap.} A larger area indicates a higher frequency of occurrence. Likewise for the figure below.}
            \label{fig:supp_correct_answer_treemap}
        \end{subfigure}\\[1ex] 
        \begin{subfigure}{\linewidth}
            \centering
            \includegraphics[width=1\linewidth]{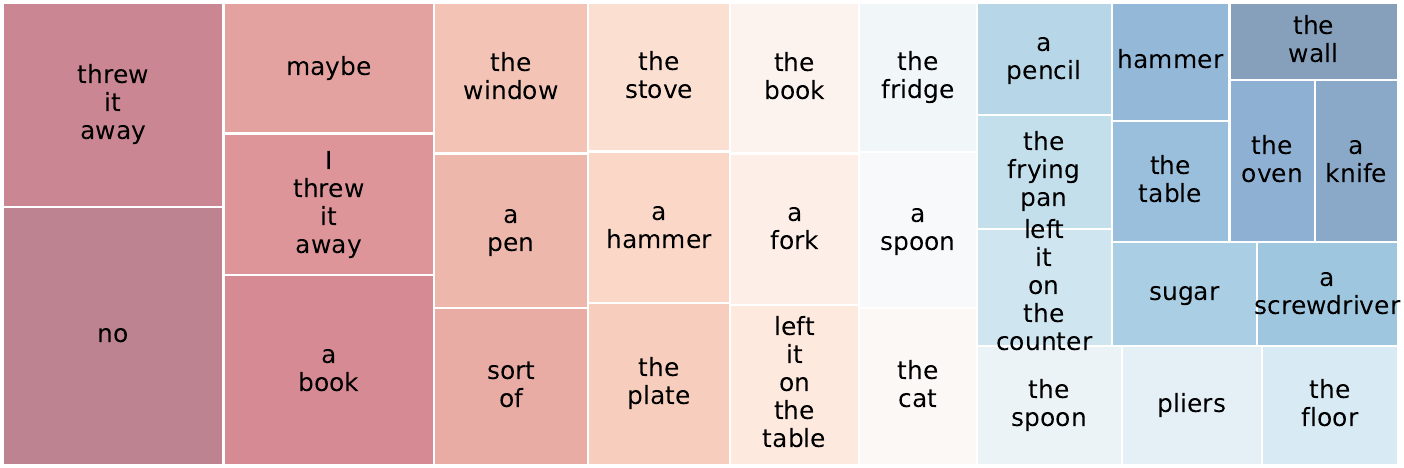}
            \caption{\textbf{Wrong answer treemap.}}
            \label{fig:supp_wrong_answer_treemap}
        \end{subfigure}
    \end{subfigure}
    \par
    \begin{subfigure}{.49\textwidth}
        \centering
        \includegraphics[width=1\linewidth]{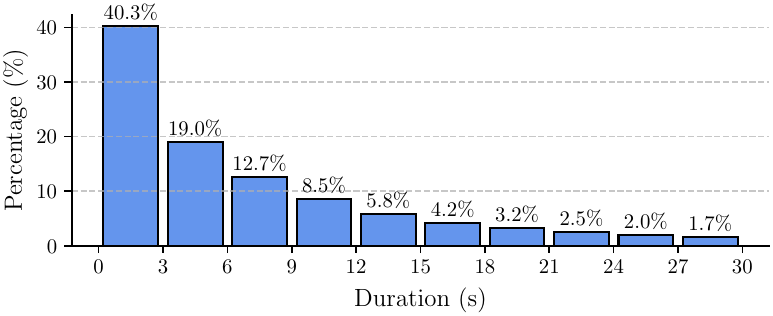}
        \caption{\textbf{Histogram of temporal windows for VLG.}}
        \label{fig:supp_duration_hist}
    \end{subfigure}%
    \hfill
    \begin{subfigure}{.49\textwidth}
        \centering
        \includegraphics[width=1\linewidth]{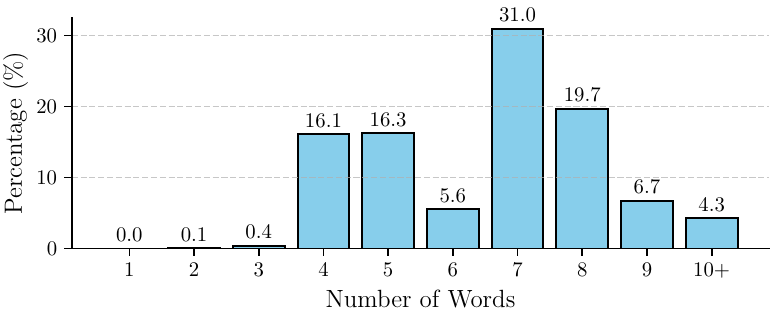}
        \caption{\textbf{Histogram of question word count.}}
        \label{fig:supp_question_word_count}
    \end{subfigure}
    \par
    \begin{subfigure}{.49\textwidth}
        \centering
        \includegraphics[width=1\linewidth]{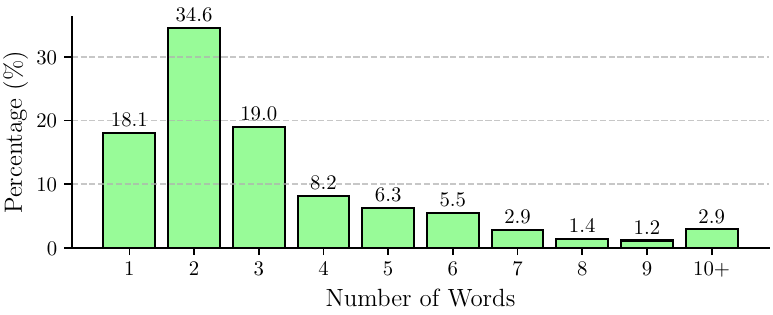}
        \caption{\textbf{Histogram of correct answer word count.}}
        \label{fig:supp_correct_answer_word_count}
    \end{subfigure}
    \hfill
    \begin{subfigure}{.49\textwidth}
        \centering
        \includegraphics[width=1\linewidth]{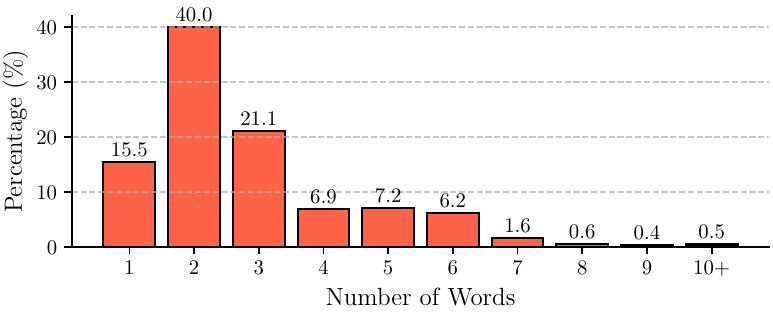}
        \caption{\textbf{Histogram of wrong answer word count.}}
        \label{fig:supp_wrong_answer_word_count}
    \end{subfigure}
\vspace{-5pt}
\caption{\textbf{Combined figure of various EgoTimeQA visualizations.}}
\end{figure}

We offer additional statistical details about our \ourdataset. 
In Fig.~\ref{fig:supp_question_sunburst}, we present the question distribution based on their first four words. Most questions begin with ``what'', inquiring about objects or actions. 
Others start with ``where'', ``did'', ``how'', {\em etc.}, showcasing their diversity. 
In Fig.~\ref{fig:supp_correct_answer_treemap} and \ref{fig:supp_wrong_answer_treemap}, we present the distribution of the top 30 correct and incorrect answers, respectively. Fig.~\ref{fig:supp_duration_hist} presents a histogram of the duration of temporal windows in \ourdataset, with the majority falling within the 0-3 second range. Fig.~\ref{fig:supp_question_word_count}, \ref{fig:supp_correct_answer_word_count}, and \ref{fig:supp_wrong_answer_word_count} shows the word count distributions for questions, correct answers, and incorrect answers, respectively. Correct and incorrect answers have similar distributions, averaging 3.2 and 2.8 words respectively, while questions, averaging 6.6 words, indicate their greater complexity.
\vspace{5pt}

\clearpage
\section{Additional Qualitative Analysis}

We conduct qualitative analysis on VLG, OpenQA, and CloseQA tasks, showcasing both success and failure scenarios. In all examples, we present results from our proposed \ourmethod, the Oracle baseline, and SimpleVQA. Each model is built upon the \texttt{Flan-T5-Base} language model. Specifically, \ourmethod~is trained concurrently on VLG, OpenQA, and CloseQA tasks with \qaego4d~and \ourdataset~data. Oracle, a variant of \ourmethod, takes only the question-related video clips as input, eliminating the need for temporal grounding. SimpleVQA* is our reproduced SimpleVQA~\cite{qaego4d} model, trained on OpenQA and CloseQA tasks with \qaego4d~data. For a detailed examination, please zoom in on the figures.

\subsection{Results on OpenQA}\label{sec:supp_openqa}

\begin{figure*}[hb]
\captionsetup{font=normalsize}
\captionsetup[sub]{font=normalsize}
\setlength{\belowcaptionskip}{0.5\baselineskip}
\centering
  \begin{subfigure}{.87\linewidth}
    \centering
    \includegraphics[width=\linewidth, page=1]{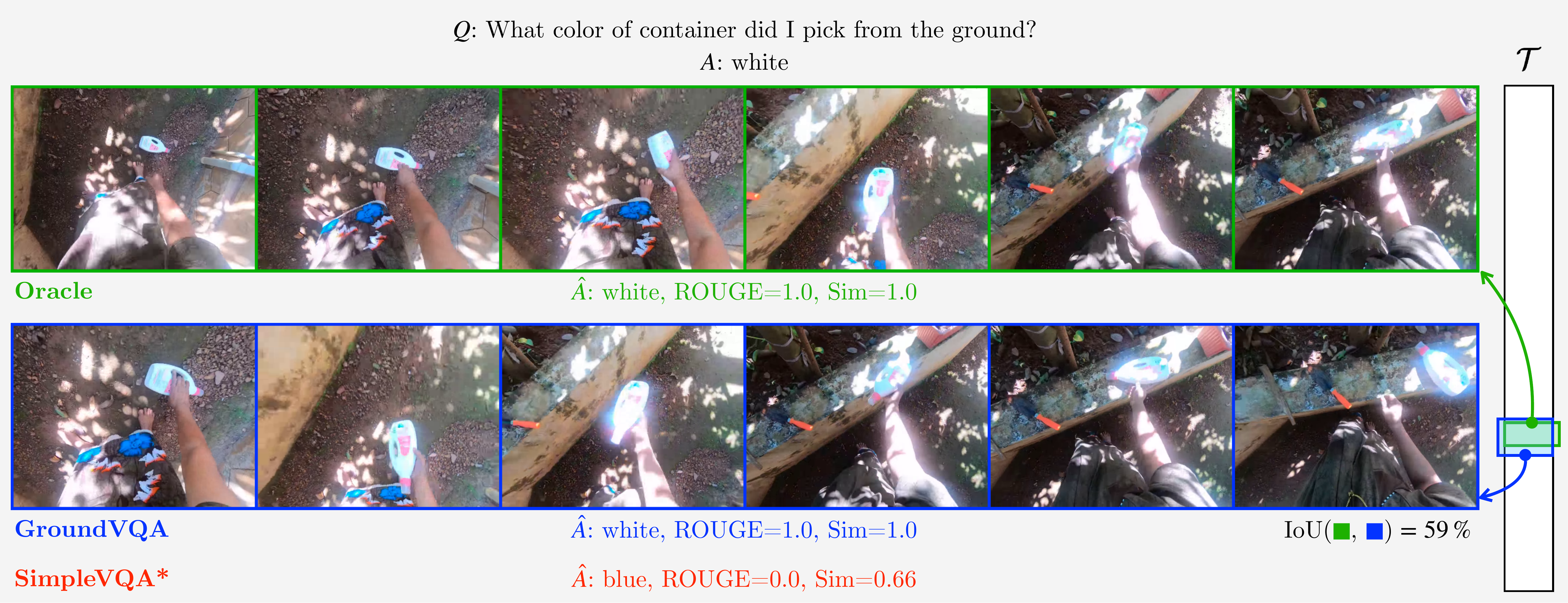}
    \caption{\textcolor{blue}{\ourmethod} has correct VLG and OpenQA predictions.}\label{fig:supp_openqa_1a}
  \end{subfigure}
  \par
  \begin{subfigure}{.87\linewidth}
    \centering
    \includegraphics[width=\linewidth, page=2]{imgs/src/supp_openqa.pdf}
    \caption{\textcolor{blue}{\ourmethod} has correct VLG and valid OpenQA predictions.}\label{fig:supp_openqa_1b}
  \end{subfigure}
\vspace{-0.4cm}
\caption{
\textbf{OpenQA success cases.}
we compare three models: \textcolor{mygreen}{Oracle}, our \textcolor{blue}{\ourmethod}, and \textcolor{red}{SimpleVQA$^*$}. 
From top to bottom are the query $\mathcal{Q}$, answer $\mathcal{A}$, six frames uniformly sampled from the grounded video segment, and the predicted answer $\hat{\mathcal{A}}$ with metrics. 
Additionally, the right side illustrates the video's time span and the predicted temporal window $\mathcal{T}$, with \textcolor{mygreen}{Oracle}'s temporal window serving as the ground truth. 
Note that SimpleVQA$^*$ is incapable of temporal grounding.
}\label{fig:supp_openqa_1}
\end{figure*}

\textbf{VLG \& OpenQA succeed.} 
In Fig.~\ref{fig:supp_openqa_1a}, \ourmethod~accurately localizes the video segment relevant to the query and correctly identifies \textit{container}'s color. Conversely, SimpleVQA* predicts a wrong color, and its lack of temporal grounding hinders error analysis. Overall, integrating the VLG task not only boosts QA performance but also enhances the interpretability of our model by clarifying the sources of errors. In Fig.~\ref{fig:supp_openqa_1b}, \ourmethod~closely predicts the temporal window and provides an answer (\textit{in the fridge}) that, while slightly different, conveys the same meaning as the ground truth (\textit{inside the refrigerator}). This case illustrates the limitation of the ROUGE metric in distinguishing between correct and incorrect paraphrased answers. Therefore, we introduce the sentence similarity metric and an additional CloseQA task to address this evaluation challenge.

\clearpage

\begin{figure*}[hb]
\captionsetup{font=normalsize}
\captionsetup[sub]{font=normalsize}
\setlength{\belowcaptionskip}{0.5\baselineskip}
\centering
  \begin{subfigure}{\linewidth}
    \includegraphics[width=\linewidth, page=3]{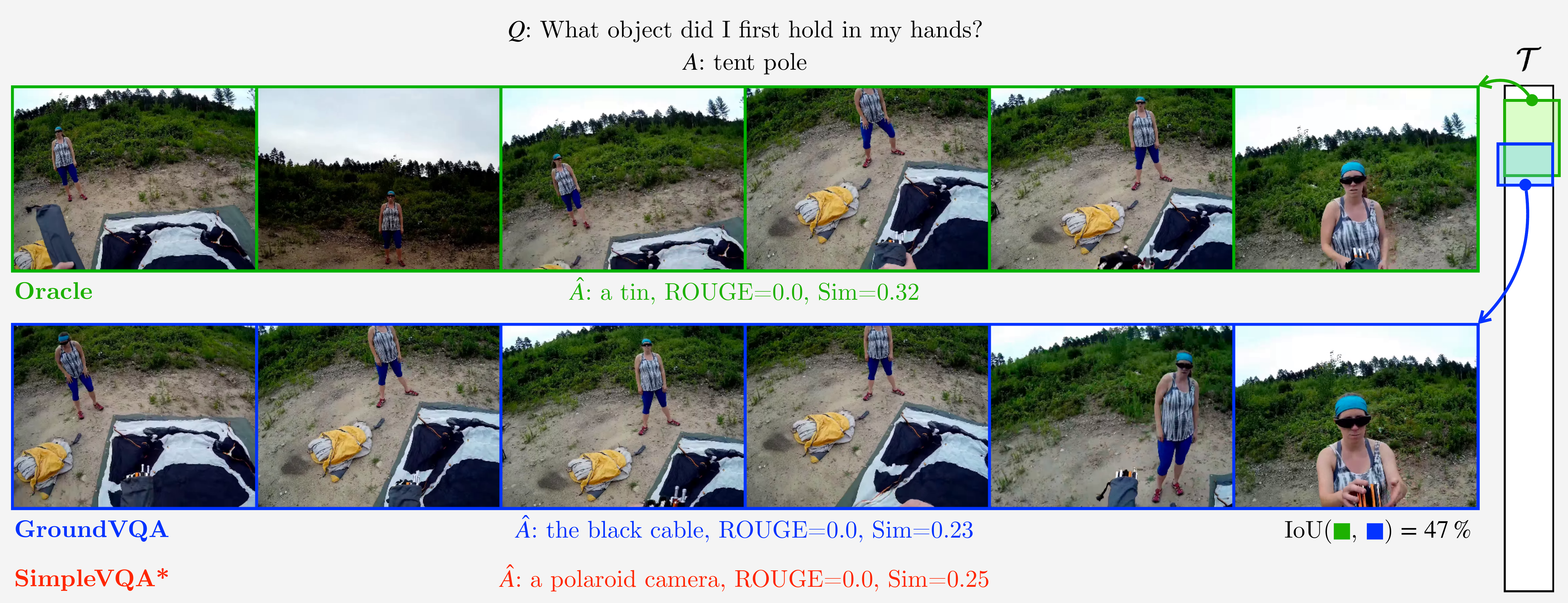}
    \caption{\textcolor{blue}\ourmethod~has correct VLG but wrong OpenQA predictions.}\label{fig:supp_openqa_2a}
  \end{subfigure}
  \par
  \begin{subfigure}{\linewidth}
    \includegraphics[width=\linewidth, page=4]{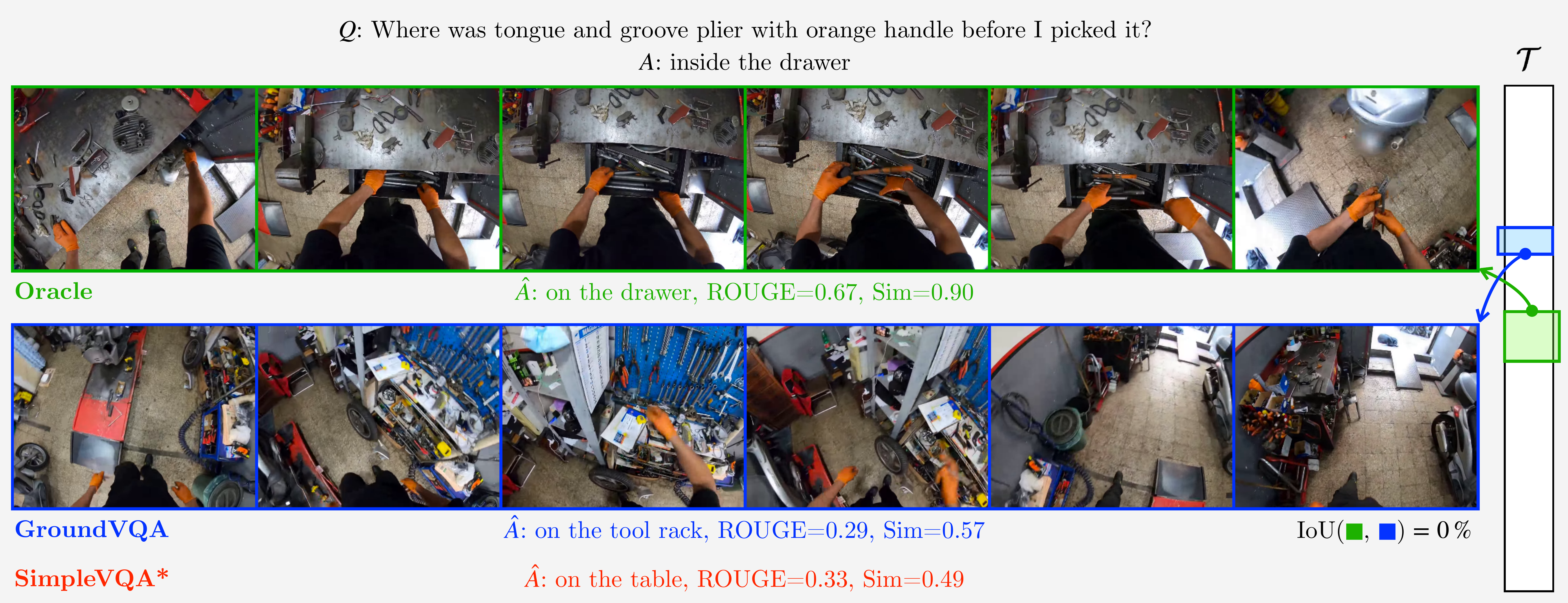}
    \caption{\textcolor{blue}\ourmethod~has wrong VLG and OpenQA predictions.}\label{fig:supp_openqa_2b}
  \end{subfigure}
\vspace{-0.6cm}
\caption{
\textbf{OpenQA failure cases.} Refer to Fig.~\ref{fig:supp_openqa_1} for the descriptions of the figures.
}\label{fig:supp_openqa_2}
\end{figure*}

\noindent\textbf{VLG succeeds, OpenQA fails.}
In Fig.~\ref{fig:supp_openqa_2a}, although \ourmethod~successfully identifies the relevant video segment, it incorrectly answers the query. The \textit{tent pole} visible in the sampled frames occupies a minor portion of the frame and is easily mistaken for similar objects, such as a \textit{tin} or \textit{black cable}, leading to errors in both Oracle and \ourmethod. In contrast, SimpleVQA*'s response (\textit{a polaroid camera}) is entirely off-topic, indicating a misdirected focus.

\vspace{10pt}
\noindent\textbf{VLG \& OpenQA fail.}
In Fig.~\ref{fig:supp_openqa_2b}, \ourmethod~fails to correctly ground the query and answer the question. SimpleVQA* also errs in its response. However, the Oracle model, with access to the ground truth video segment, provides a valid answer. A closer look at the frames grounded by \ourmethod~reveals its attention to the \textit{tongue and groove plier with an orange handle} on the tool rack, but it overlooks the action of \textit{picking} mentioned in the query. This indicates that \ourmethod~still has limitations in comprehending questions and reasoning about video content.
\vspace{20pt}

\clearpage
\subsection{Results on CloseQA}\label{sec:supp_closeqa}

\begin{figure*}[hb]
\captionsetup{font=normalsize}
\captionsetup[sub]{font=normalsize}
\setlength{\belowcaptionskip}{0.5\baselineskip}
\centering
  \begin{subfigure}{\linewidth}
    \includegraphics[width=\linewidth, page=1]{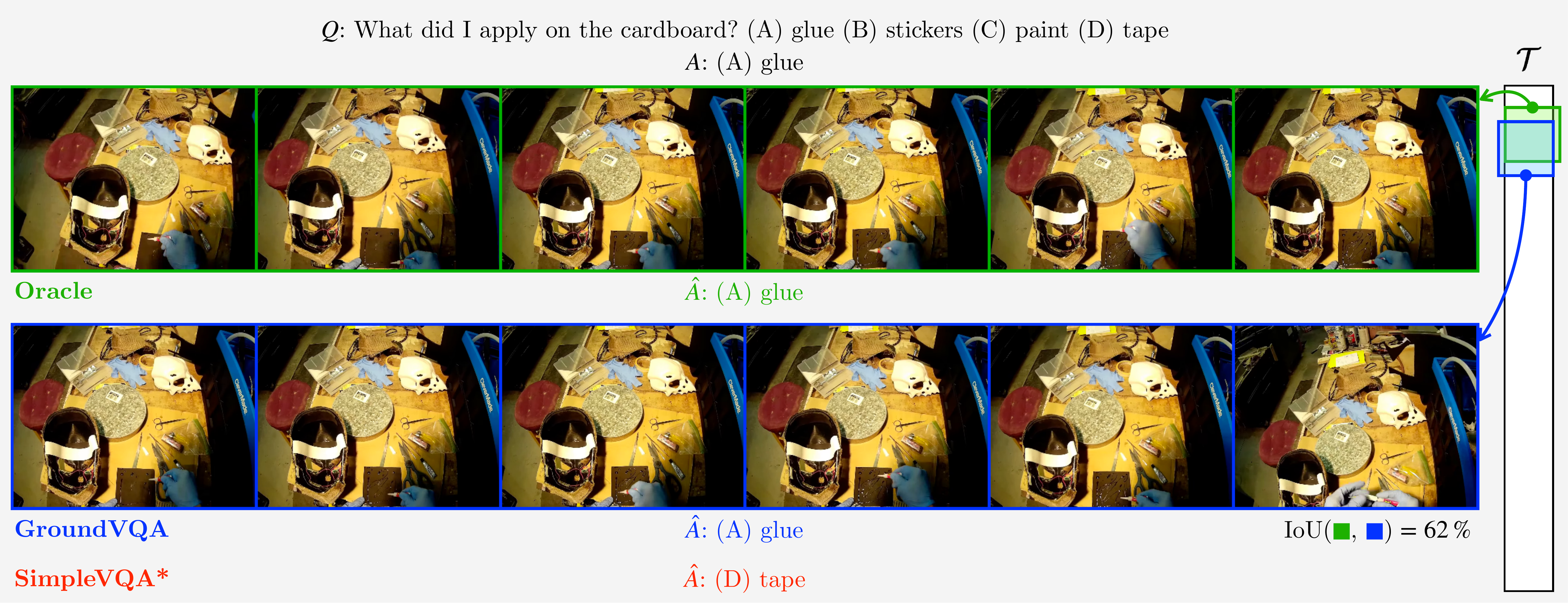}
    \caption{\textcolor{blue}{\ourmethod} has correct VLG and CloseQA predictions for a video with complex environments.}\label{fig:supp_closeqa_1a}
  \end{subfigure}
  \par
  \begin{subfigure}{\linewidth}
    \includegraphics[width=\linewidth, page=2]{imgs/src/supp_closeqa.pdf}
    \caption{\textcolor{blue}{\ourmethod} has correct VLG and CloseQA predictions.}\label{fig:supp_closeqa_1b}
  \end{subfigure}
\vspace{-0.6cm}
\caption{
\textbf{CloseQA success cases.} Refer to Fig.~\ref{fig:supp_openqa_1} for the descriptions of the figures.
}\label{fig:supp_closeqa_1}
\end{figure*}

\noindent\textbf{VLG \& CloseQA succeed.}
In Fig.~\ref{fig:supp_closeqa_1a}, \ourmethod~successfully localizes the video clip corresponding to the question and selects the correct answer. Notice how tiny the \textit{glue} is in the grounded video frames, which demonstrates our method's object recognition capability. On the contrary, SimpleVQA* chooses an incorrect option, and the reason for this is unclear. This example also highlights the advantage of the CloseQA task, which eliminates ambiguities and paraphrasing dilemmas in evaluating open-ended answers.

In Fig.~\ref{fig:supp_closeqa_1b}, \ourmethod~excels in both temporal grounding and question-answering, whereas SimpleVQA* fails. This example underscores the importance and challenge of temporal grounding, as the model needs to identify the \textit{carton} target and recognize the \textit{open} action. Once the precise temporal window is grounded, identifying the \textit{small knife} used to \textit{open the carton} becomes significantly easier.
\vspace{20pt}

\clearpage

\begin{figure*}[hb]
\captionsetup{font=normalsize}
\captionsetup[sub]{font=normalsize}
\setlength{\belowcaptionskip}{0.5\baselineskip}
\centering
  \begin{subfigure}{\linewidth}
    \includegraphics[width=\linewidth, page=3]{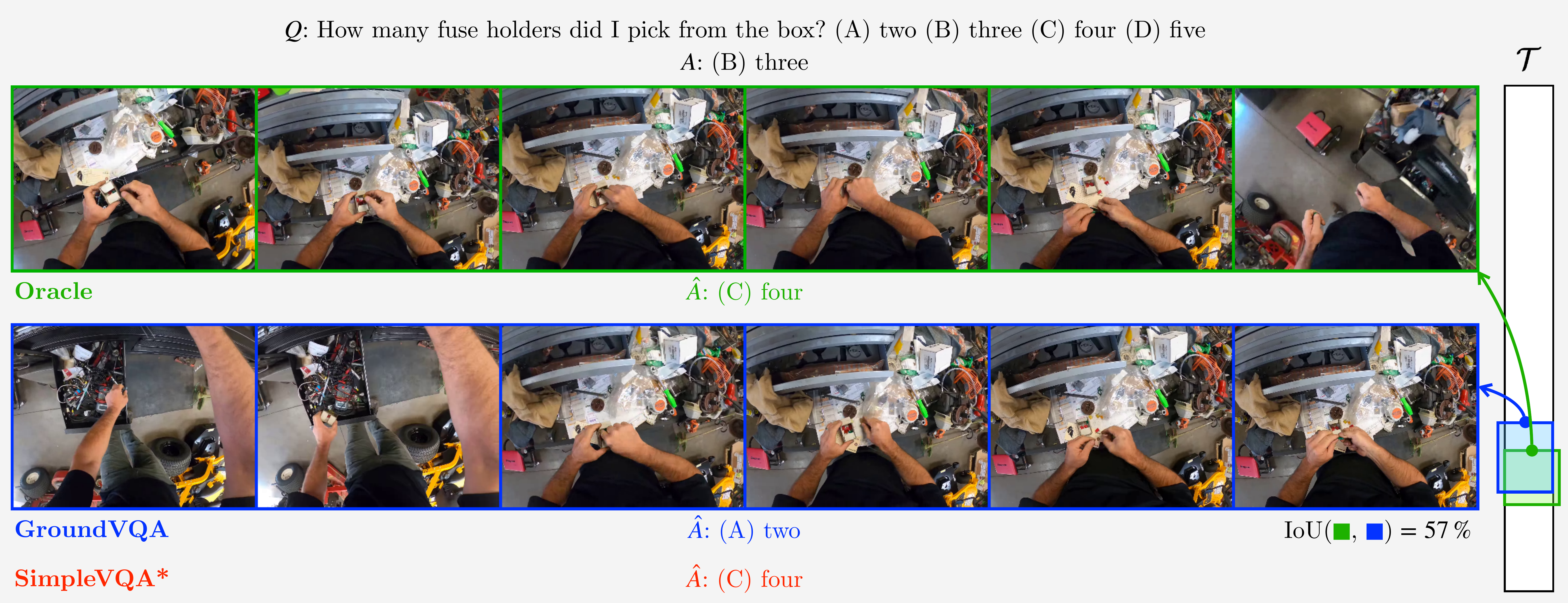}
    \caption{\textcolor{blue}\ourmethod~has correct VLG but wrong CloseQA predictions.}\label{fig:supp_closeqa_2a} 
  \end{subfigure}
  \par
  \begin{subfigure}{\linewidth}
    \includegraphics[width=\linewidth, page=4]{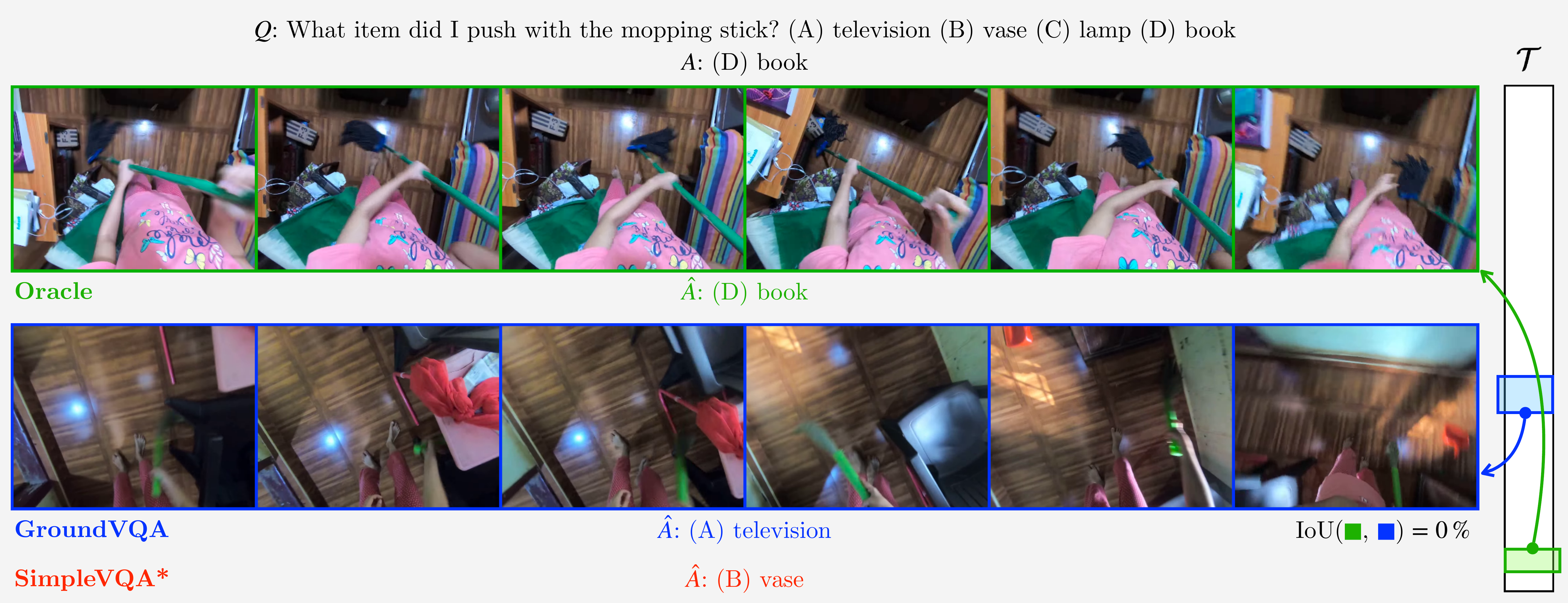}
    \caption{\textcolor{blue}\ourmethod~has wrong VLG and CloseQA predictions.}\label{fig:supp_closeqa_2b}
  \end{subfigure}
\vspace{-0.6cm}
\caption{
\textbf{CloseQA failure cases.} Refer to Fig.~\ref{fig:supp_openqa_1} for the descriptions of the figures.
}\label{fig:supp_closeqa_2}
\end{figure*}

\noindent\textbf{VLG succeeds, CloseQA fails.}
In Fig.~\ref{fig:supp_closeqa_2a}, \ourmethod~succeeds in temporal grounding. However, all three models choose the wrong answer. This example, which involves a counting task, highlights the models' limitations in counting objects across sequential frames. Future improvements could include training with more data, incorporating object-centric representations~\cite{locatello2020object,elsayed2022savi++,zhang2023helping}, or adopting object detection/tracking techniques~\cite{tokmakov2021learning,li2023video}.

\vspace{10pt}
\noindent\textbf{VLG \& CloseQA fail.}
In Fig.~\ref{fig:supp_closeqa_2b}, \ourmethod~fails in both query grounding and answering. SimpleVQA* also fails. By contrast, the Oracle model, with access to relevant video content, chooses the correct answer. The frames grounded by \ourmethod~contains the \textit{mopping stick} but miss the \textit{pushing} action and the \textit{television} it erroneously selects.
\vspace{20pt}

\clearpage

\section{Full Prompts for Data Generation} \label{sec:supp_prompt}

Here, we list the full prompts utilized for generating OpenQA and CloseQA data. Text in~\texttt{\textcolor{red}{red}} indicates variable inputs.

\subsection{Generate OpenQA Data Using Llama-2}
\vspace{-3pt}

\begin{lstlisting}
<s>[INST] <<SYS>>
You are an AI Assistant and always write the output of your response in JSON. I will provide you with a series of narrations that depict my behavior. You should generate one QA pair based on the narrations in the format of {"Q": <question>, "A": <answer>}. In the narrations, "C" represents me, and "O" represents someone else. Use as much information as possible from narrations to generate the question, and the question you generate should be able to be answered using the information provided in the narrations. The question should be in the past tense. The question should be within 10 words, and the answer should be within 5 words. <</SYS>>

C pours hot water from the frying pan in his left hand into the bowl in his right hand. [/INST] {"Q": "What did I pour in the bowl?", "A": "boiling water"} </s>
<s>[INST] C searches through the cabinet. C closes the cabinet. C picks the tin from the cabinet. C places the tin on the counter. [/INST] {"Q": "Where was the tin before I took it?", "A": "at the cabinet"} </s>
<s>[INST] C turns on sink knob. C washes the cucumber on the sink. C turns off sink knob. [/INST] {"Q": "Did I wash the cucumber?", "A": "yes"} </s>
<s>[INST] @<narrations>@ [/INST]
\end{lstlisting}

\subsection{Generate OpenQA Data Using ChatGPT}
\vspace{-3pt}

\begin{lstlisting}
You're an AI Assistant, outputting responses in JSON. I'll give behavior narrations, in which "C" is me, "O" is someone else. Generate a QA pair like {"Q": <question>, "A": <answer>} based on them. The question should use the narration info, be in the past tense, <= 10 words, and the answer <= 5 words.

User: C pours hot water from the frying pan in his left hand into the bowl in his right hand
Assistant: {"Q": "What did I pour in the bowl?", "A": "boiling water"}

User: C searches through the cabinet. C closes the cabinet. C picks the tin from the cabinet. C places the tin on the counter.
Assistant: {"Q": "Where was the tin before I took it?", "A": "at the cabinet"}

User: C turns on sink knob. C washes the cucumber on the sink. C turns off sink knob.
Assistant: {"Q": "Did I wash the cucumber?", "A": "yes"}

User: @<narrations>@
\end{lstlisting}

\subsection{Generate CloseQA Data Using Llama-2}
\vspace{-3pt}

\begin{lstlisting}
<s>[INST] <<SYS>>
I'll provide a question and its correct answer. Generate three plausible, but incorrect, answers that closely resemble the correct one Make it challenging to identify the right answer. No preamble, get right to the three wrong answers and present them in a list format. <</SYS>>

Question: How many frying pans can i see on the shelf? Correct Answer: two pieces. Wrong Answers: [/INST] ["one piece", "three piece", "five pieces"] </s>
<s>[INST] Question: What colour bowl did i carry from the plate stand? Correct Answer: green. Wrong Answers: [/INST] ["blue", "black", "white"] </s>
<s>[INST] Question: What did i pour in the bowl? Correct Answer: boiling water. Wrong Answers: [/INST] ["hot oil", "steamed milk", "warm broth"] </s>
<s>[INST] Question: @<question>@ Correct Answer: @<answer>@. Wrong Answers: [/INST]
\end{lstlisting}

\section{Limitations and Future Work}
From the experiments and analysis presented in the main paper and this supplementary material, we can draw the following observations.
{\em First}, the performance of our method is closely tied to the quality of video features and training data. Enhancements in video features, particularly through improved visual-language alignment, and the inclusion of more training data, could lead to future improvements.
{\em Second}, despite our efforts in designing appropriate prompts and rigorously filtering data, biases and inaccuracies still exist in the LLM-generated data.
{\em Third}, our method faces challenges in fine-grained perception tasks, for example, object recognition and counting, particularly in complex environments. The adoption of object-centric features, for example, those for tracking and counting techniques could enhance performance in these areas.
{\em Fourth}, processing long egocentric videos demands significant computational resources. Future research should explore the use of memory networks for compressing video features and developing more efficient models that maintain accuracy.
{\em Finally}, a query may relate to multiple video segments, but Ego4D-NLQ and QA-Ego4D assume only one is relevant per question. We advocate for loosening this assumption, as it is typical for multiple personal experiences to be triggered by a single query. Considering the difficulty of labeling multiple temporal windows for one query, a practical solution is to employ a well-trained Vision-Language Model (VLM) to identify potential candidates for further confirmation via human review. Subsequently, our method can be applied to generate QA pairs for each confirmed candidate.

\vspace{2em}
{
    \small
    \bibliographystyle{ieeenat_fullname}
    \bibliography{main}
}

\end{document}